\documentclass{article} 
\usepackage{iclr2019_conference,times}

\usepackage{amsmath,amsfonts,bm}

\def\eqref#1{(\ref{#1})}

\def\1{\bm{1}}

\DeclareMathAlphabet{\mathsfit}{\encodingdefault}{\sfdefault}{m}{sl}
\SetMathAlphabet{\mathsfit}{bold}{\encodingdefault}{\sfdefault}{bx}{n}

\DeclareMathOperator*{\argmin}{arg\,min}

\usepackage{hyperref}
\usepackage{url}

\usepackage[utf8]{inputenc} 
\usepackage[T1]{fontenc}    
\usepackage{booktabs}       
\usepackage{amsfonts,amsmath}       
\usepackage{nicefrac}       
\usepackage{multirow}
\usepackage{graphicx}
\usepackage{makecell}
\usepackage[para,online,flushleft]{threeparttable}
\PassOptionsToPackage{numbers, compress}{natbib}
\usepackage{verbatim}
\usepackage{wrapfig}

\usepackage{algorithm}
\usepackage{algpseudocode}
\usepackage[small,bf]{caption}
\usepackage{subcaption}
\usepackage{mathtools}

\usepackage{lipsum}
\usepackage{color}
\definecolor{Sijia_color}{rgb}{0.858, 0.188, 0.478}
\definecolor{Kd_color}{rgb}{0.5, 0.388, 0.878}
\definecolor{Pu_color}{rgb}{0.1, 0.8, 0.8}
\definecolor{Xue_color}{rgb}{0, 0, 1}
\definecolor{PY_color}{rgb}{0.2, 0.8, 0}
\definecolor{QF}{rgb}{0.8, 0.2, 0.5}

\newtheorem{myprop}{\bf{Proposition}}

\DeclarePairedDelimiterX{\inp}[2]{\langle}{\rangle}{#1, #2}

\DeclareMathOperator*{\minimize}{\text{minimize}}

\DeclareMathOperator*{\st}{\text{subject to}}
\DeclareMathAlphabet\mathbfcal{OMS}{cmsy}{b}{n}
\newcommand{\Def}[0]{\mathrel{\mathop:}=}

\usepackage{adjustbox}

\title{Structured Adversarial Attack: \\
Towards General Implementation and Better Interpretability
}

\author{Kaidi Xu$^{1}$\thanks{Equal contribution}
\quad Sijia Liu$^{2*}$~~~Pu Zhao$^{1}$~~~Pin-Yu Chen$^{2}$~~~Huan Zhang$^{3}$~~~Quanfu Fan$^{2}$\\
\textbf{Deniz Erdogmus}$^{1}$~~~\textbf{Yanzhi Wang}$^{1}$~~~\textbf{Xue Lin}$^{1}$\\
  $^1$Northeastern University, USA\\
    $^2$MIT-IBM Watson AI Lab, IBM Research, USA\\
    $^3$University of California, Los Angeles, USA  
}

\iclrfinalcopy 

\begin{document}

\maketitle

\vspace*{-0.1in}
\begin{abstract}
\vspace*{-0.1in}
When generating adversarial examples to attack deep neural networks (DNNs), $\ell_p$ norm of the added perturbation is usually used to measure the similarity between original image and adversarial example. However, such adversarial attacks \textcolor{black}{perturbing the raw input spaces} may fail to capture \textcolor{black}{structural} information hidden in the input. This work develops a more general attack model, i.e., the structured attack \textcolor{black}{(StrAttack)}, which explores group sparsity in adversarial perturbations by sliding a mask through images aiming for extracting key \textcolor{black}{spatial}  structures. An ADMM (alternating direction method of multipliers)-based framework is proposed that can split the original problem into a sequence of analytically solvable subproblems and can be generalized to implement other attacking methods. Strong group sparsity  is achieved in adversarial perturbations even with the same level of
$\ell_p$-norm distortion ($p \in \{1,2,\infty \}$) as the state-of-the-art attacks. We demonstrate  the effectiveness of \textcolor{black}{StrAttack} by extensive experimental results on MNIST, CIFAR-10 and ImageNet. We also show that StrAttack provides better interpretability (i.e., better correspondence with discriminative image regions) through adversarial saliency map \citep{papernot2016limitations} and class activation map \citep{zhou2016learning}.

\end{abstract}

\section{Introduction}

\begin{wrapfigure}{r}{0.55\textwidth} 
\centering
\hspace*{-0.6cm}
\includegraphics[width=0.9\linewidth]{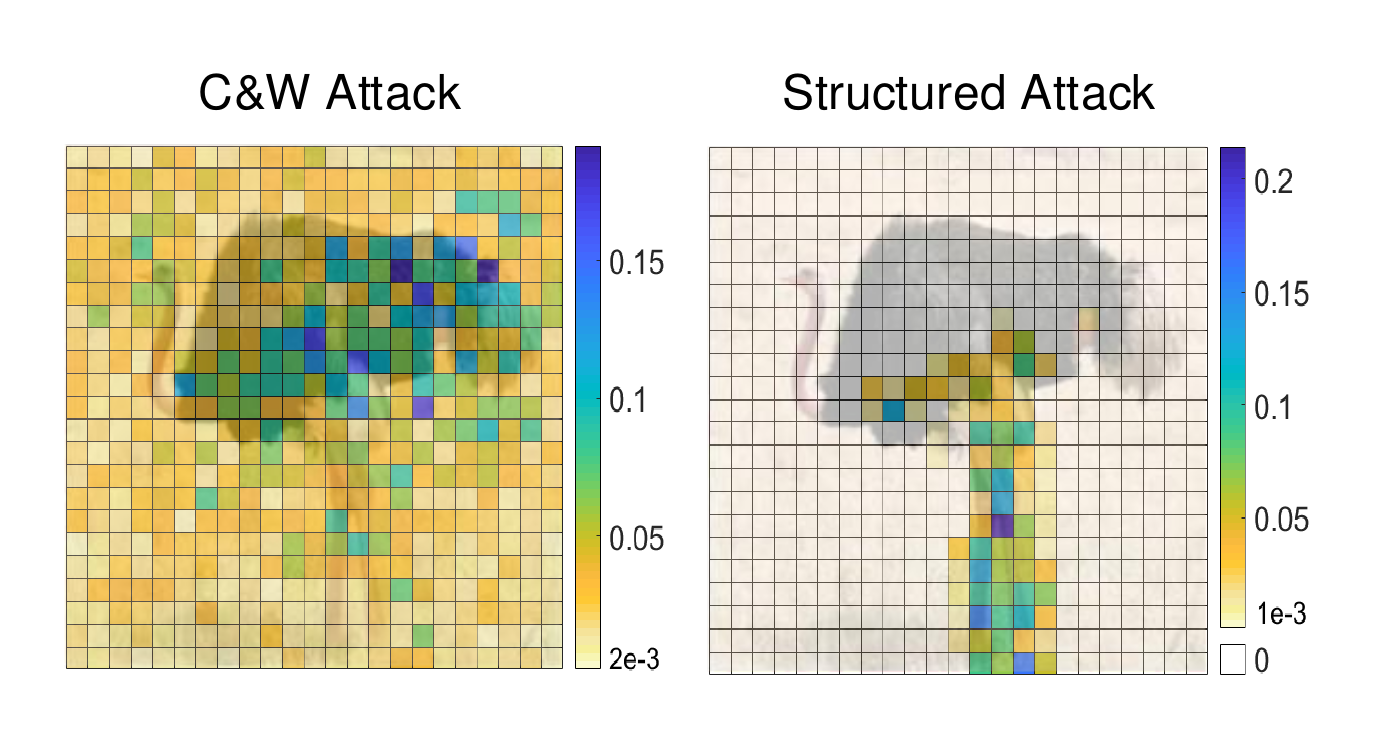}
\vspace*{-0.2cm}
\caption{Group sparsity demonstrated in adversarial perturbations obtained by  C\&W attack and our StrAttack, 
where `ostrich' is the original label, and `unicycle' is the mis-classified label. Here each group is a region of $ 13 \times 13 \times 3$  pixels  and the strength of adversarial perturbations (through their $\ell_2$ norm) at each group  is represented by heatmap. C\&W attack perturbs almost all groups, while StrAttack 
yields strong group sparsity, with more  semantic  structure: 
the perturbed image region matches the  feature of the target object, namely, the frame of the unicycle. 
}
\label{fig:show}
\vspace{-0.2cm}
\end{wrapfigure}

Deep learning 
achieves exceptional successes in domains such as image recognition \citep{he2016deep,geifman2017selective}, natural language processing \citep{hinton2012deep,harwath2016unsupervised}, medical diagnostics \citep{chen2016combining,SHI201814} and advanced control \citep{silver2016mastering,fu2017ex2}. 
Recent studies \citep{szegedy2013intriguing,goodfellow2014explaining,nguyen2015deep,kurakin2016adversarial,carlini2017towards} show that DNNs are vulnerable to adversarial attacks implemented by generating adversarial examples, i.e., adding well-designed perturbations to original legal inputs. Delicately crafted adversarial examples can mislead a DNN to recognize them as any target \textcolor{black}{image label},
while \textcolor{black}{the perturbations appears unnoticeable to human eyes.} 
Adversarial attacks against DNNs not only exist in theoretical models but also pose potential security threats to the real world \citep{kurakin2016adversarial,evtimov2017robust,papernot2017practical}.
Several explanations are proposed to illustrate why there exist adversarial examples to DNNs based on hypotheses such as model linearity and data manifold \citep{goodfellow2014explaining,gilmer2018adversarial}. However, little is known to their origins, and convincing explanations remain to be explored.


Besides achieving \textcolor{black}{the goal of (targeted)} mis-classification, an adversarial example should be as ``similar'' to the original legal input as possible to be stealthy.
Currently, the similarity is measured by the $\ell_p$ norm $(p=0,1,2,\infty)$ of the added perturbation \citep{szegedy2013intriguing,carlini2017towards,chen2017zoo,chen2017ead}, i.e., $\ell_p$ norm is being minimized when generating adversarial example.
However, measuring the similarity between the original image and its adversarial example by $\ell_p$ norm is neither necessary nor sufficient \citep{mahmood2018onthesuitability}. Besides, no single measure can be perfect for human perceptual similarity \citep{carlini2017towards} and such adversarial attacks may fail to capture key information hidden in the input such as spatial structure or distribution. 
Spurred by that, this work implements a new attack model i.e., \emph{structured attack (StrAttack)} that imposes group sparsity
on adversarial perturbations by extracting structures from the inputs.
As shown in Fig.\,\ref{fig:show},  we find that StrAttack  identifies minimally sufficient regions that make  attacks successful, but without incurring extra pixel-level perturbation power.
The major contributions are summarized as below.
\begin{itemize}
\item (\textbf{Structure-driven attack})  This work is the first attempt towards exploring group-wise sparse structures when implementing adversarial attacks, but without losing $\ell_p$ distortion performance when compared to state-of-the-art attacking methods.
\item (\textbf{Generality}) We show that 
the proposed  attack model  covers many norm-ball based attacks  such as C\&W \citep{carlini2017towards} and EAD \citep{chen2017ead}.
\item (\textbf{Efficient implementation})
We develop an efficient algorithm to generate structured adversarial perturbations by leveraging \textcolor{black}{the} alternating direction method of multipliers (ADMM). We show that ADMM splits the original complex problem into subproblems, each of which can be solved \textit{analytically}. Besides, we show that ADMM can further be used to refine an arbitrary adversarial attack under the fixed sparse structure.
\item (\textbf{Interpretability})
The generated adversarial perturbations demonstrate clear correlations and interpretations between original  and target images. With the aid of adversarial saliency map \citep{papernot2016limitations} and class activation map \citep{zhou2016learning}, we show that the obtained group-sparse adversarial patterns better shed light on the mechanisms of adversarial perturbations to fool DNNs.
\end{itemize}

\paragraph{Related work}
\textcolor{black}{
 Many works  studied  norm-ball  constrained  adversarial  attacks. For example, 
FGM 
\citep{goodfellow2014explaining} and IFGSM  
\citep{KurakinGB2016adversarial} attack methods were proposed to maximize the classification error 
subject to $\ell_\infty$-norm based distortion constraints. Moreover,
  L-BFGS  \citep{szegedy2013intriguing} and C\&W \citep{carlini2017towards} attacks found an adversarial example by
  minimizing its $\ell_2$-norm distortion.
By contrast, JSMA \citep{papernot2016limitations} and one-pixel \citep{su2017one} attacks attempted to generate adversarial examples by perturbing the minimum number of pixels, namely,  minimizing the $\ell_0$ norm of adversarial perturbations. 
 Different from the above norm-ball constrained attacks, some works \citep{karmon2018lavan,brown2017adversarial} crafted adversarial examples by adding noise patches. However, the resulting adversarial perturbations are no longer imperceptible  to humans. Here we argue that imperceptibility could be important since it helps us to understand how/why DNNs are vulnerable to adversarial attacks while
  perturbing natural examples just by  indistinguished adversarial noise.
}

\textcolor{black}{In the aforementioned  norm-ball  constrained  adversarial  attacks,  two extremely opposite principles have been applied: C\&W attack (or  $\ell_\infty$ attacks) seeks the minimum image-level  distortion but allows to modify all pixels; 
one-pixel attack only perturbs a few pixels but suffers a high pixel-level distortion.
Both attacking principles  might lead to a high noise visibility due to perturbing too many pixels or perturbing a few pixels too much. In this work, we wonder if there exists a more effective attack that can be as successful as existing attacks but achieves a tradeoff between the  perturbation power and the number of perturbed pixels. We will show that the proposed StrAttack is able to identify sparse perturbed regions that make attacks successful, but without incurring extra pixel-level perturbations. 
 It is also worth mentioning that one-pixel attack has much lower attack success rate on ImageNet than C\&W attack and StrAttack.
} 


\textcolor{black}{
In addition to adversarial attacks, many defense works have been proposed. Examples include    
defensive distillation \citep{papernot2016distillation}  that distills the original DNN and  introduces temperature into the softmax layer, random mask \citep{anonymous2019random} that modifies the DNN structures by randomly  removing certain neurons before training,
\textcolor{black}{adversarial  training through enlarging the training dataset with adversarial examples,}
and robust adversarial  training  \citep{madry2017towards,sinha2018certifying} through the min-max optimization. 
It is commonly known that the robust adversarial training method ensures the strongest defense performance against adversarial attacks on MNIST and CIFAR-10. 
In this work, we will evaluate the effectiveness of StrAttack to three defense methods, a) defensive distillation \citep{papernot2016distillation}, b) adversarial training via data augmentation \citep{tram2018ensemble} and c) robust adversarial training \citep{madry2017towards}.
}

\textcolor{black}{
Although the adversarial attack  and   defense  have attracted an increasing amount of attention,
the visual explanation on  adversarial perturbations is less explored since the distortion power is minimized and
the resulting adversarial effects become imperceptible to humans.
The work \citep{dong2017towards} attempted to understand how the internal representations of DNNs are affected by adversarial examples. However, only an ensemble-based attack was considered, which fails to distinguish the effectiveness of different norm-ball constrained adversarial attacks.
Unlike  \citep{dong2017towards}, we employ the interpretability tools, adversarial  saliency map  (ASM) \citep{papernot2016limitations} and class activation map (CAM) \citep{zhou2016learning} to measure the effectiveness of different attacks in terms of their interpretability. 
Here ASM provides sensitivity analysis for
  pixel-level perturbation's impact on label classification, and CAM  localizes class-specific   image discriminative regions \citep{DBLP:journals/corr/abs-1801-02612}.
We will show that the  sparse adversarial pattern obtained by StrAttack offers a great interpretability through ASM  and  CAM compared with other norm-ball constrained attacks.
}

\section{Structured Attack: Explore group structures from images}

\label{sec: concept_group}

In the section, we introduce the concept of \textit{StrAttack},
motivated by the question: `what possible structures could adversarial perturbations have to fool DNNs?' Our idea is to  divide an image into sub-groups of pixels and then penalize  the corresponding group-wise sparsity. The resulting sparse groups encode minimally sufficient adversarial effects on local structures of natural images.

Let $\boldsymbol{\Delta} \in \mathbb R^{W \times H \times C}$ be an adversarial perturbation added to an original image $\mathbf X_0$, where $W \times H$ gives the spatial region, and $C$ is the depth, e.g., $C = 3$ for RGB images. To characterize the   local structures   of $\boldsymbol{\Delta}$, we introduce a \textit{sliding mask} $\mathcal M$   with stride $S$ and size $r \times r \times C$.  When $S = 1$,  the mask moves one pixel at a time; When $S = 2$, the mask jumps $2$ pixels at a time while sliding.
By
  adjusting the stride $S$ and the mask size $r$, different group splitting schemes can be obtained. If $S < r$, the resulting groups will contain \textit{overlapping} pixels. By contrast, groups will become \textit{non-overlapped} when $S = r$. 
 
 A sliding mask $\mathcal M$ finally divides $\boldsymbol{\Delta}$ into a set of groups $ \{ \boldsymbol{\Delta}_{\mathcal G_{p,q}} \}  $ 
for $p \in [P]$ and $q \in [Q]$, where $P = (W-r)/S + 1$, $Q = (H-r)/S+1$, and $[n]$ denotes the integer set $\{1,2,\ldots,n \}$. 
 Given the groups $ \{ \boldsymbol{\Delta}_{\mathcal G_{p,q}} \}  $, 
 the group sparsity can be {characterized through the following sparsity-inducing function \citep{yuan2006model,bach2012optimization,liu2015sparsity}, motivated by the problem of group Lasso \citep{yuan2006model}:}
 \begin{align}\label{eq: group_Delta}
 \small
\begin{array}{c}
     g(\boldsymbol{\Delta}) = 
     \sum_{p=1}^P \sum_{q=1}^Q \| \boldsymbol{\Delta}_{\mathcal G_{p,q}} \|_{2},
\end{array}
      %
 \end{align}
where $\boldsymbol{\Delta}_{\mathcal G_{p,q}} $ denotes the set of pixels  of $\boldsymbol{\Delta}$  indexed by $\mathcal G_{p,q}$, and $\|\cdot \|_2$ is the $\ell_2$ norm.
We refer readers to Fig.\,\ref{fig: overlap} for an illustrative example of  our concepts on groups and group sparsity. 

\section{Structured Adversarial Attack with ADMM}\label{sec: structure_ADMM}

In this section, we
start by proposing a general framework to generate  prediction-evasive adversarial examples, where 
the  adversary   relies only on  gradients of the loss function with
respect to   inputs of DNNs.
Our model takes into account both  commonly-used adversarial distortion metrics   and the proposed   group-sparsity regularization that encodes spatial structures in attacks. We show that the   process of  generating structured adversarial examples leads to an  optimization problem that is difficult to solve using the existing optimizers Adam (for C\&W attack) and  FISTA (for EAD attack) \citep{carlini2017towards,chen2017ead}. 
To circumvent this challenge,  we develop an efficient optimization method   via   alternating direction method of multipliers (ADMM).

Given an original image $\mathbf x_0 \in \mathbb R^n$, we aim to design the optimal adversarial perturbation $\boldsymbol{\delta} \in \mathbb R^n$ so that the adversarial example $(\mathbf x_0 + \boldsymbol{\delta})$ misleads DNNs trained on natural images. 
{Throughout this paper, we use   vector representations of the adversarial perturbation $\boldsymbol{\Delta}$ and the original image 
$\mathbf X_0$ without loss of generality.}
A well designed perturbation $\boldsymbol \delta$ can be obtained by solving optimization problems of the following form,
\begin{align}\label{eq: prob}
\small
\begin{array}{ll}
    \displaystyle \minimize_{\boldsymbol \delta } & f(\mathbf x_0 + \boldsymbol \delta, t) + \gamma D(\boldsymbol \delta) + \tau 
    g(\boldsymbol{\delta})
    \\
  \st    &  (\mathbf x_0 + \boldsymbol \delta) \in [0,1]^n, ~ \| \boldsymbol \delta \|_\infty \leq \epsilon,
\end{array}
\end{align}
where $f(\mathbf x, t)$ denotes the loss function for   crafting adversarial
example given a target class   $t$,
$D(\boldsymbol{\delta })$ is a distortion function that controls \textcolor{black}{the} perceptual similarity between a natural image and a perturbed image,
$ g(\boldsymbol{\delta}) = \sum_{p=1}^{P} \sum_{q=1}^{Q} \| \boldsymbol{\delta}_{\mathcal G_{p,q}}\|_2$ is given by \eqref{eq: group_Delta},
and
 $\| \cdot \|_p$ signifies  the $\ell_p$ norm.
In problem \eqref{eq: prob},
the `hard' constraints ensure the validness of created adversarial examples with $\epsilon$-tolerant perturbed pixel values.  And the non-negative regularization parameters $\gamma$ and $\tau$ place our emphasis on the  distortion  of an adversarial example (to an original image) and group sparsity of adversarial perturbation. 
{Tuning the regularization parameters will be discussed in
\textcolor{black}{Appendix\,\ref{parameter_setting}}.
}

Problem \eqref{eq: prob} 
gives a quite
general formulation for design of adversarial examples.
If we remove the group-sparsity regularizer $g(\boldsymbol{\delta})$ and  the $\ell_\infty$ constraint, problem \eqref{eq: prob} becomes the same as the C\&W attack \citep{carlini2017towards}. More specifically, if we further set the distortion function $D(\boldsymbol \delta)$ to the form of $\ell_0$, $\ell_2$ or $\ell_\infty$ norm,
then we obtain   C\&W $\ell_0$, $\ell_2$ or $\ell_\infty$ attack. If  $D(\boldsymbol \delta)$ is specified by the elastic-net regularizer, then  problem
\eqref{eq: prob} becomes the formulation of  
 EAD attack \citep{chen2017ead}.

In this paper, 
 we specify the loss function of problem \eqref{eq: prob} as below, which   yields the best known performance of adversaries  \citep{carlini2017towards},
\begin{align}\label{eq: fx}
\small
f(\mathbf x_0 + \boldsymbol \delta, t) = c \cdot \max \{ \max_{j \neq t}  Z(\mathbf x_0+ \boldsymbol \delta)_j -Z(\mathbf x_0+ \boldsymbol \delta )_t, - \kappa \},
\end{align}
where $Z(\mathbf x)_j$ is the $j$th element of logits $Z(\mathbf x)$, representing the output before the last softmax layer in DNNs, and $\kappa$ is a confidence parameter
that is usually set to zero if  the attack transferability is not much cared. 
We   choose   $D(\boldsymbol{\delta }) = \|\boldsymbol{\delta }  \|_2^2 $ for a fair comparison with  the { C\&W $\ell_2$ adversarial attack}. In this section, we   assume that
$\{ \mathcal G_{p,q} \}$ are non-overlapping groups, i.e., $\mathcal G_{p,q} \cap \mathcal G_{p^\prime,q^\prime} = \emptyset $ for $q \neq q^\prime$ or $p \neq p^\prime$. 
The overlapping case will be studied in the next section.

 
The presence of \textit{multiple} non-smooth regularizers and `hard' constraints make the existing optimizers Adam and FISTA  \citep{carlini2017towards,chen2017ead,KingmaB2015adam,beck2009fast} inefficient for solving problem \eqref{eq: prob}. First, the subgradient of the objective function of problem \eqref{eq: prob} is difficult to obtain especially when   $\{\mathcal G_{p,q} \}$ are overlapping groups.  Second,  it is impossible to compute the proximal
operations required for FISTA with respect to  all non-smooth regularizers and `hard' constraints. Different from the existing work, we show that ADMM, 
a first-order   operator splitting method, helps us to split the original complex problem \eqref{eq: prob} into a sequence of subproblems, each of which can be solved  \textit{analytically}.

We reformulate problem \eqref{eq: prob}
in a way that lends itself to the application of ADMM,
\begin{align}\label{eq: prob_ADMM_non}
\small
\begin{array}{ll}
    \displaystyle \minimize_{\boldsymbol \delta, \mathbf z, \mathbf w, \mathbf y} & f(\mathbf z + \mathbf x_0) + \gamma D(\boldsymbol \delta) + \tau \sum_{i=1}^{PQ} \| \mathbf y_{\mathcal D_i}\|_2 + h(\mathbf w) \\
  \st    
  & \mathbf z =  \boldsymbol \delta,~     \mathbf z = \mathbf y, ~ \mathbf z = \mathbf w,
\end{array}
\end{align}
where $\mathbf z$, $\mathbf y$ and $\mathbf w$ are newly introduced variables, for ease of notation let $\mathcal D_{(q-1)P + p} = \mathcal G_{p,q}$, and $h(\mathbf w)$ is an indicator function with respect to the constraints of problem \eqref{eq: prob},
\begin{equation}\label{eq: indicator}
\small
    h(\mathbf w) = \left \{
    \begin{array}{ll}
        0 &  \text{if $(\mathbf x_0 + \mathbf w) \in [0,1]^n, ~ \| \mathbf w \|_\infty \leq \epsilon$,} \\
      \infty   & \text{otherwise}.
    \end{array}
    \right.
\end{equation}
ADMM is performed by minimizing the augmented Lagrangian of problem \eqref{eq: prob_ADMM_non},
\begin{align}
\small
\begin{array}{c}
 L( \mathbf z, \boldsymbol \delta,  \mathbf y, \mathbf w, \mathbf u, \mathbf v, \mathbf s) = f( \mathbf z + \mathbf x_0 ) + \gamma
  D(\boldsymbol \delta) + \tau  \textstyle \sum_{i=1}^{PQ} \| \mathbf y_{\mathcal D_i}\|_2 + h(\mathbf w) + \mathbf u^T (\boldsymbol \delta - \mathbf z)   \\
  \hspace*{0.5in} + \mathbf v^T (\mathbf y -  \mathbf z)+ \mathbf s^T (\mathbf w - \mathbf z)+ \frac{\rho}{2} \| \boldsymbol \delta - \mathbf z\|_2^2 + \frac{\rho}{2} \| \mathbf y -  \mathbf z \|_2^2 + \frac{\rho}{2} \| \mathbf w - \mathbf z\|_2^2,
\end{array}
   \label{eq: aug_Lag_non}
\end{align}
where  $\mathbf u$, $\mathbf v$ and $\mathbf s$  are Lagrangian multipliers, and $\rho > 0$ is a given penalty   parameter. 
ADMM   splits all of optimization variables into \textit{two} blocks and 
adopts   the following iterative scheme,
{\small\begin{align}
  & \{ \boldsymbol \delta^{k+1}, \mathbf w^{k+1}, \mathbf y^{k+1} \} = \argmin_{\boldsymbol{\delta},\mathbf w,\mathbf y} L(\boldsymbol \delta, \mathbf z^k, \mathbf w, \mathbf y, \mathbf u^k, \mathbf v^k, \mathbf s^k), \label{eq: delta_w_step}\\
  &   \mathbf z^{k+1}= \argmin_{\mathbf z} L(\boldsymbol \delta^{k+1}, \mathbf z, \mathbf w^{k+1}, \mathbf y^{k+1}, \mathbf u^k, \mathbf v^k, \mathbf s^k), \label{eq: z_step} \\
  & \left \{
  \begin{array}{l}
        \mathbf u^{k+1} = \mathbf u^k + \rho (\boldsymbol \delta^{k+1} - \mathbf z^{k+1}), \\
         \mathbf v^{k+1} = \mathbf v^k + \rho (\mathbf y^{k+1} - \mathbf z^{k+1}), \\
        \mathbf s^{k+1} = \mathbf s^k + \rho (\mathbf w^{k+1} - \mathbf z^{k+1}) ,
  \end{array}
  \right.
  \label{eq: dual_update}
\end{align}}%
where $k$ is the iteration index,
steps \eqref{eq: delta_w_step}-\eqref{eq: z_step} are used for updating primal variables, and the  last step \eqref{eq: dual_update} is known as the dual update step. We emphasize that the crucial property of the proposed ADMM
approach  is that, as we demonstrate in Proposition\,\ref{prop: prop1}, the solution to
  problem \eqref{eq: delta_w_step}  can be found   in parallel and exactly.


\begin{myprop}\label{prop: prop1}
When $D(\boldsymbol{\delta}) = \| \boldsymbol{\delta}\|_2^2$, the solution to problem \eqref{eq: delta_w_step} is given by
\begin{align}
&  
\begin{array}{c}
\small
\boldsymbol \delta^{k+1} = \frac{\rho}{\rho + 2\gamma}
\mathbf a,
\end{array}
\label{eq: delta_sol} \\
&  
\begin{array}{c}
\small
[\mathbf w^{k+1}]_i = \left \{
    \begin{array}{ll}
       \min \{ 1-[\mathbf x_0]_i, \epsilon \}  &  b_i >  \min \{ 1-[\mathbf x_0]_i, \epsilon \} \\
     \max \{ - [\mathbf x_0]_i, -\epsilon \}   &  b_i <  \max \{ - [\mathbf x_0]_i, -\epsilon \}  \\
     b_i & \text{otherwise},
    \end{array}
    \right. ~ \text{~for~} i \in [n],
\end{array}
    \label{eq: w_sol} \\
    &
    \begin{array}{c}
    \small
[ \mathbf y^{k+1} ]_{\mathcal D_i} = \left (
1 - \frac{\tau}{\rho \|  [ \mathbf c ]_{\mathcal D_i} \|_2}
\right )_+ [  \mathbf c ]_{\mathcal D_i},~  i \in [PQ],
    \end{array}
 \label{eq: y_sol}
\end{align}
where $\mathbf a \Def \mathbf z^k -  \mathbf u^k/\rho$, $\mathbf b \Def \mathbf z^k  -  \mathbf s^k / \rho $, $\mathbf c \Def  \mathbf z^k -   \mathbf v^k / \rho $, $(x)_+ = x$ if $x \geq 0$ and $0$ otherwise, $[\mathbf x]_i$ denotes the $i$th element of $\mathbf x$, and $[\mathbf x]_{\mathcal D_i}$ denotes the sub-vector of $\mathbf x$ indexed by $\mathcal D_i$.
\end{myprop}
\textbf{Proof:} See Appendix\,\ref{proof_Pro_1}.  
\hfill $\square$


It is clear from Proposition\,\ref{prop: prop1} that introducing auxiliary variables 
does not  increase the computational complexity of ADMM since \eqref{eq: delta_sol}-\eqref{eq: y_sol} can be solved in parallel. Moreover, 
if another    distortion metric (different from  $D(\boldsymbol{\delta}) = \| \boldsymbol{\delta}\|_2^2$) is used, then  ADMM only changes at the $\boldsymbol{\delta}$-step \eqref{eq: delta_sol}.   

We next   focus on the $\mathbf z$-minimization step \eqref{eq: z_step}, which can be equivalently transformed into
\begin{align}\label{eq: prob_z_complete_non}
\small
    \begin{array}{ll}
\displaystyle \minimize_{\mathbf z}         &  \displaystyle f(\mathbf x_0 + \mathbf z) + \frac{\rho}{2} \|    \mathbf z - \mathbf a^\prime \|_2^2 
+
\frac{\rho}{2}
\|    \mathbf z - \mathbf b^\prime \|_2^2 +   \frac{\rho}{2} \| \mathbf z -  \mathbf c^\prime \|_2^2,
    \end{array}
\end{align}
where
$
\mathbf a^\prime \Def \boldsymbol{\delta}^{k+1} +   \mathbf u^k / \rho
$, 
$
\mathbf b^\prime \Def \mathbf w^{k+1}  +   \mathbf s^k / \rho
$,
and $\mathbf c^\prime \Def \mathbf y^{k+1} +   \mathbf v^k/\rho
$. We recall that  attacks studied in this paper belongs to `first-order' adversaries  \citep{madry2017towards}, which only have access to gradients of the loss function $f$. Spurred by that, we solve problem \eqref{eq: prob_z_complete_non} via a linearization technique that is commonly used in stochastic/online ADMM \citep{ouyang2013stochastic,suzuki2013dual,liu2017zeroth} or linearized ADMM  \citep{boyd2011distributed,liu2017linearized}.
Specifically,
we replace the function $f$ with its first-order Taylor expansion  at the point $\mathbf z^k$ by adding a Bregman divergence term $(\eta_k/2 )\| \mathbf z - \mathbf z^k \|_2^2$.
As a result, problem \eqref{eq: prob_z_complete_non}  becomes 
\begin{align}
\small
    \begin{array}{ll}
\displaystyle \minimize_{\mathbf z}         &  \displaystyle (\nabla f(\mathbf z^k + \mathbf x_0))^T ( \mathbf z  - \mathbf z^k) + \frac{\eta_k}{2} \| \mathbf z  - \mathbf z^k \|_2^2 + \frac{\rho}{2} \|  \mathbf z - \mathbf a^\prime \|_2^2  \\
&\displaystyle +
\frac{\rho}{2}
\|   \mathbf z- \mathbf b^\prime \|_2^2 +   \frac{\rho}{2} \| \mathbf z - \mathbf c^\prime \|_2^2,
    \end{array}
    \label{eq: prob_theta_complete_linear}
\end{align}
where $1/\eta_k > 0$ is a given decaying parameter, e.g., $\eta_k = \alpha \sqrt{k}$ for some $\alpha > 0$, and the Bregman divergence term stabilizes the convergence of $\mathbf z$-minimization step.
It is clear that
 problem \eqref{eq: prob_theta_complete_linear} yields a quadratic program with the closed-form solution
\begin{equation}\label{eq: z_sol_non}
\small
    \mathbf z^{k+1} = \left (1/ \left ( \eta_k + 3 \rho \right ) \right)
    \left ( \eta_k \mathbf z^k + \rho \mathbf a + \rho \mathbf b +\rho     \mathbf c- \nabla f(\mathbf z^k + \mathbf x_0)
    \right ).
\end{equation}

In summary, the proposed ADMM algorithm alternatively updates 
\eqref{eq: delta_w_step}-\eqref{eq: dual_update},   which yield closed-form solutions given by  \eqref{eq: delta_sol}-\eqref{eq: y_sol} and \eqref{eq: z_sol_non}.  
The convergence of linearized ADMM for nonconvex optimization 
was recently   proved by \citep{liu2017linearized}, and thus provides   theoretical validity of our approach. Compared to the existing solver for generation of adversarial examples \citep{carlini2017towards,papernot2016limitations}, our algorithm offers two main benefits, \textit{efficiency} and  \textit{generality}. That is, the computations for every update step are efficiently carried out, and our approach can be applicable to a wide class
of attack formulations.

\section{Overlapping Group and Refined StrAttack
}

In this section, we generalize our proposed ADMM solution framework to the case of generating adversarial perturbations with \textit{overlapping} group structures. We then turn to \textcolor{black}{an attack refining} model under \textit{fixed} sparse structures. We will show that both extensions can be unified under the ADMM framework. In particular, the refined approach will allow us to gain deeper insights on the structural effects on adversarial perturbations.

\subsection{Overlapping group structure}\label{sec: overlap_ADMM}

We recall    that groups $\{ \mathcal D_i \}$ (also denoted by $\{ \mathcal G_{p,q} \}$) studied in Sec.\,\ref{sec: structure_ADMM}
could be overlapped with each other; {see an example in Fig.\,\ref{fig: overlap}.}
Therefore, $\{ \mathcal D_i \}$  is in general  a \textit{cover} rather than a {partition} of $[n]$.
 To address the challenge in coupled group variables, we introduce multiple copies of   the variable $\mathbf y$ in problem \eqref{eq: prob_ADMM_non}, and achieve the following modification
 \begin{align}\label{eq: prob_ADMM_overlap}
 \small
\begin{array}{ll}
    \displaystyle \minimize_{\boldsymbol \delta, \mathbf z, \mathbf w, \{ \mathbf y_i \}} & f(\mathbf z + \mathbf x_0) + \gamma D(\boldsymbol \delta) + \tau \sum_{i=1}^{PQ} \| \mathbf y_{i,\mathcal D_i}\|_2 + h(\mathbf w) \\
  \st    
  & \mathbf z =  \boldsymbol \delta, ~ \mathbf z = \mathbf w,~     \mathbf z = \mathbf y_i,\quad i \in [PQ],
\end{array}
\end{align}%
where compared to problem \eqref{eq: prob_ADMM_non}, there exist $PQ$ variables $\mathbf y_i \in \mathbb R^n$ for $i \in [PQ]$, and $\mathbf y_{i,\mathcal D_i}$ denotes the subvector of $\mathbf y_i$ with indices given by $\mathcal D_i$. It is clear from  \eqref{eq: prob_ADMM_overlap} that groups $\{  \mathcal D_i \}$ become \textit{non-overlapped} since each of them lies in a different copy $\mathbf y_i$. 
The ADMM algorithm for solving problem \eqref{eq: prob_ADMM_overlap} maintains a similar procedure  as
\eqref{eq: delta_w_step}-\eqref{eq: dual_update} except $\mathbf y$-step  \eqref{eq: y_sol} and $\mathbf z$-step \eqref{eq: z_sol_non}; see 
Proposition\,\ref{prop: prop2}.

\begin{myprop}\label{prop: prop2}
Given the same condition of Proposition\,\ref{prop: prop1}, the ADMM solution to problem \eqref{eq: prob_ADMM_overlap} involves the   $\boldsymbol{\delta}$-step  same as \eqref{eq: delta_sol}, the   $\mathbf w$-step same as \eqref{eq: w_sol}, and two modified $\mathbf y$- and $\mathbf z$-steps,
{\small \begin{align}
&
\begin{array}{c}
\left \{
\begin{array}{l}
 \left [ \mathbf y_i^{k+1} \right  ]_{\mathcal D_i} =      \left (
 1 - \frac{\tau}{\rho \|  [ \mathbf c_i ]_{\mathcal D_i} \|_2}
 \right )_+ \left [  \mathbf c_i \right ]_{\mathcal D_i} \\
    \left [ \mathbf y_i^{k+1} \right ]_{[n]/\mathcal D_i} = \left [ \mathbf c_i \right ]_{[n]/\mathcal D_i}
\end{array}
\right., ~\text{for}~ i \in [PQ],
\end{array}
\label{eq: y_sol_overlap}\\
& 
\begin{array}{c}
\small 
\mathbf z^{k+1} =  
\left(1/\left(
\eta_k + 2\rho +PQ \rho
\right )
\right ) \left (
 \eta_k \mathbf z^k + \rho \mathbf a^\prime + \rho \mathbf b^\prime +\rho  \sum_{i=1}^{PQ}   \mathbf c_i^\prime- \nabla f(\mathbf z^k + \mathbf x_0) 
\right ),
\end{array}
\label{eq: z_sol_overlap}
\end{align} }%
where $\mathbf c_i \Def \mathbf z^k - \mathbf v_i^k/\rho$, $\mathbf v_i$ is the Lagrangian multiplier associated with equality constraint $\mathbf y_i = \mathbf z $, similar to  \eqref{eq: dual_update} we obtain $\mathbf v_i^{k+1} = \mathbf v_i^k + \rho (\mathbf y^{k+1} - \mathbf z^{k+1})$,
$[n]/\mathcal D_i$ denotes the difference of sets $[n]$ and $\mathcal D_i$, $\mathbf a^\prime$ and $\mathbf b^\prime$ have been defined in \eqref{eq: prob_z_complete_non}, and $\mathbf c_i^\prime = \mathbf y_i^{k+1} + \mathbf v_i^k/\rho$.
\end{myprop}
\textbf{Proof}:  {See Appendix\,\ref{proof_Pro_2}. 
} 
\hfill $\square$

We note that updating $PQ$  variables $\{ \mathbf y_i \}$ is decomposed as shown in  \eqref{eq: y_sol_overlap}. However, the side effect is the need of $PQ$ times more storage space than the $\mathbf y$-step \eqref{eq: y_sol} when   groups are non-overlapped.

\subsection{Refined StrAttack under fixed sparse pattern
}\label{refined}
The approaches proposed in Sec.\,\ref{sec: structure_ADMM} and Sec.\,\ref{sec: overlap_ADMM} help us to identify  structured sparse patterns  in adversarial perturbations. 
This section presents a method to refine  structured attacks under fixed group sparse patterns. 
Let 
{$\boldsymbol{\delta}^*$} denote 
the  solution to problem \eqref{eq: prob} solved by the proposed ADMM method.  We   define a   $\sigma$-sparse perturbation $\boldsymbol{\delta}$ via $\boldsymbol{\delta}^*$,
{\small \begin{align}\label{eq: sig_sparse}
    \delta_i = 0~ \text{~if~} \delta_i^* \leq \sigma,~  \text{for any } i \in [n],
\end{align}}%
where a hard thresholding operator is applied to $\boldsymbol{\delta}^*$ with tolerance $\sigma$. 
Our refined model imposes
 the fixed $\sigma$-sparse structure \eqref{eq: sig_sparse} into problem \eqref{eq: prob}. This leads to
{\small \begin{align}\label{eq: prob_retrain}
\begin{array}{ll}
    \displaystyle \minimize_{\boldsymbol \delta } & f(\mathbf x_0 + \boldsymbol \delta) + \gamma D(\boldsymbol \delta) \\
  \st    &  (\mathbf x_0 + \boldsymbol \delta) \in [0,1]^n, ~ \| \boldsymbol \delta \|_\infty \leq \epsilon \\
  & \delta_i = 0, ~\text{if}~ i \in \mathcal S_\sigma, 
\end{array}
\end{align}}%
where $\mathcal  S_\sigma$ is defined by \eqref{eq: sig_sparse}, i.e., $\mathcal S_\sigma \Def \{ j\, | \, \delta_j^* \leq \sigma,\, j \in [n] \}$. Compared to 
problem \eqref{eq: prob}, the group-sparse penalty function is eliminated as it has been known as \textit{a priori}. With the priori knowledge of group sparsity, problem \eqref{eq: prob_retrain} is formulated to optimize and refine the non-zero groups, thus achieving better performance on highlighting and exploring the perturbation structure. 
Problem \eqref{eq: prob_retrain} can be solved using ADMM, and its solution is presented in Proposition\,\ref{prop: prop3}.

\begin{myprop}\label{prop: prop3}
The ADMM solution to problem \eqref{eq: prob_retrain} is given by 
{\small\begin{align}
  &  [\boldsymbol{\delta}^{k+1}]_i = \left \{
    \begin{array}{ll}
     0 & i \in \mathcal S_\sigma \\
       \min \{ 1- \left [\mathbf x_0 \right ]_i, \epsilon \}  &  
       \frac{\rho}{2\gamma+\rho} a_i >  \min \{ 1- \left [ \mathbf x_0  \right ]_i, \epsilon \},~ i \notin \mathcal S_\sigma \\
     \max \{ -  \left [ \mathbf x_0  \right ]_i, -\epsilon \}   &  
     \frac{\rho}{2\gamma+\rho} a_i <  \max \{ -  \left [ \mathbf x_0  \right ]_i, -\epsilon \}, i \notin \mathcal S_\sigma  \\
        \frac{\rho}{2\gamma+\rho} a_i & \text{otherwise},
    \end{array}
    \right. \\
    & \displaystyle [\mathbf z^{k+1}]_i = \left \{
     \begin{array}{ll}
        0  &  i \in \mathcal S_\sigma \\
        1/(\eta_k +   \rho) \left [ \eta_k [\mathbf z^k]_i + \rho [\mathbf a^\prime]_i - [\nabla f(\mathbf z^k + \mathbf x_0)]_i \right ] & i \notin \mathcal S_\sigma,
     \end{array}
     \right.
\end{align}}
for $i \in [n]$, where $\mathbf z = \boldsymbol{\delta}$ is the   introduced auxiliary variable similar to \eqref{eq: prob_ADMM_non},   $\mathbf a \Def \boldsymbol{\delta}^{k+1} - \mathbf u^k/\rho$,
$\mathbf a^\prime \Def \boldsymbol{\delta}^{k+1} + \mathbf u^k/\rho$, $\mathbf u^{k+1} = \mathbf u^k + \rho (\boldsymbol{\delta}^{k+1} - \mathbf z^{k+1})$, and $\rho $ and $\eta_k$ have been defined in \eqref{eq: aug_Lag_non} and \eqref{eq: prob_theta_complete_linear}.
The ADMM iterations can be initialized by $\boldsymbol{\delta}^*$, the known solution to problem  \eqref{eq: prob}. 
\end{myprop}
\textbf{Proof}:
See Appendix\,\ref{supp: prop3}. 
\hfill $\square$

\vspace*{-0.1in} 
\section{Empirical Performance of StrAttack 
}
\vspace*{-0.1in}

We evaluate the performance of the proposed StrAttack
on three image classification datasets, MNIST \citep{Lecun1998gradient}, CIFAR-10 \citep{Krizhevsky2009learning} and ImageNet \citep{deng2009imagenet}.
To make fair comparison with the C\&W $\ell_2$ attack \citep{carlini2017towards}, 
we use $\ell_2$ norm as the distortion function $D(\boldsymbol{\delta}) = \| \boldsymbol{\delta}\|_2^2$. And we also compare with FGM \citep{goodfellow2014explaining} and IFGSM $\ell_2$ attacks \citep{KurakinGB2016adversarial} as a reference.
We evaluate attack success rate (ASR)\footnote{The percentage of adversarial examples that successfully fool DNNs.} as well as $\ell_p$ distortion metrics for $p \in \{0,1,2,\infty \}$. 
\textcolor{black}{The detailed experiment setup is presented in Appendix\,\ref{parameter_setting}. 
Our code is available at \url{https://github.com/KaidiXu/StrAttack}.
}


For each attack method on MNIST or CIFAR-10, we choose 1000 original images from the test dataset as source and each image has 9 target labels. So a total of 9000 adversarial examples are generated for each attack method. On ImageNet, each attack method tries to craft 900 adverdarial examples with 100 random images from the test dataset and 9 random target labels for each image.

\begin{figure}[htb]  
  \centering
  \vspace*{-0.15in}
\includegraphics[width=.95\textwidth]{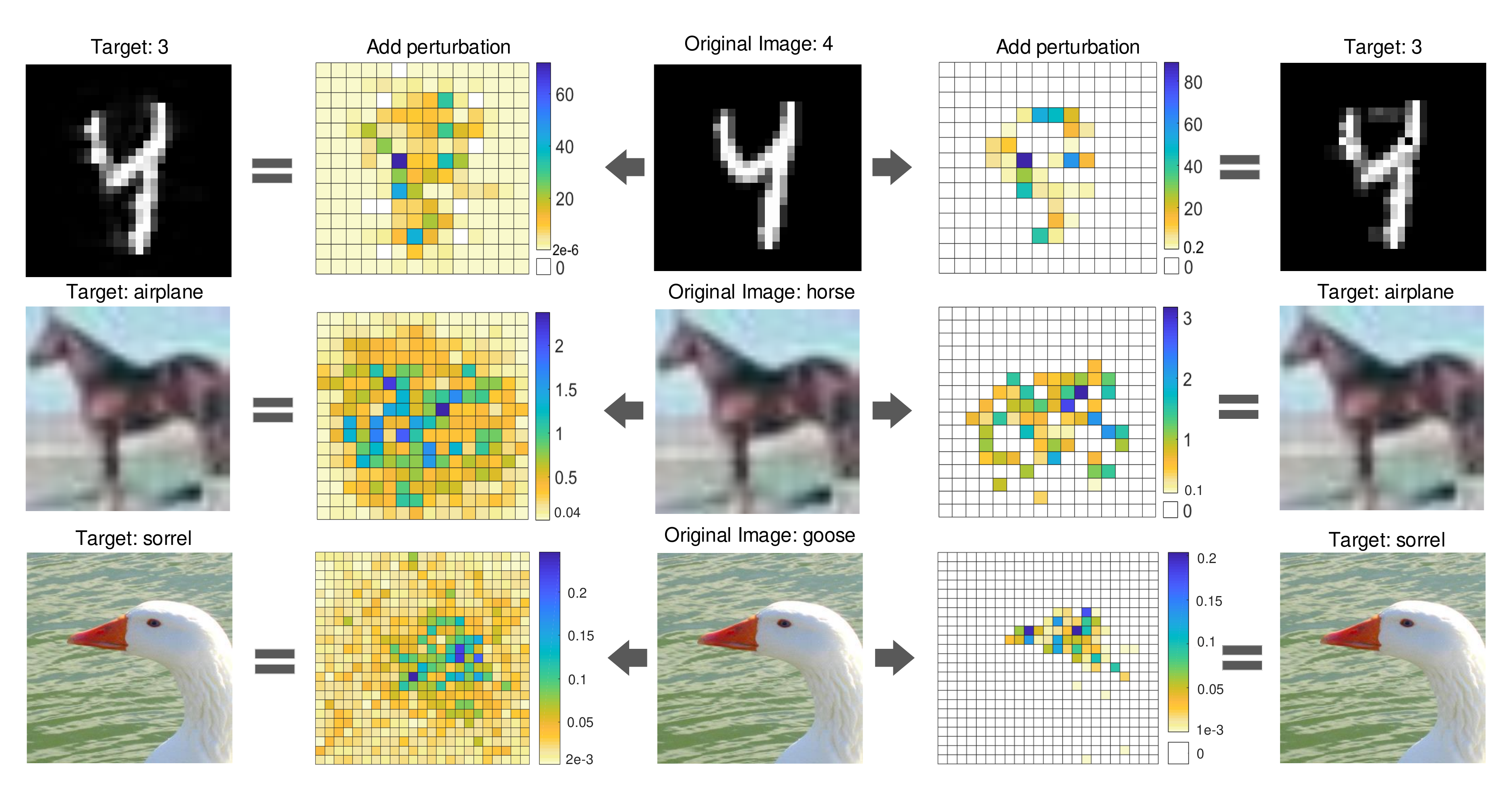}
\vspace*{-0.1in}
\caption{{C\&W attack vs StrAttack. Here each grid cell represents a  $2 \times 2$, $2 \times 2$, and $13 \times 13$ small region in  MNIST, CIFAR-10 and ImageNet, respectively. The group sparsity of perturbation is represented by heatmap. The colors on heatmap represent average absolute value of distortion scale to $[0,255]$. The left two columns correspond to results of using C\&W attack. The right two columns show results of StrAttack.}
}
\label{fig: samples}

\end{figure}

Fig.\,\ref{fig: samples} compares adversarial examples generated by 
StrAttack
and  C\&W attack on each dataset. 
We observe that
the perturbation of the C\&W attack has poor group sparsity, i.e., many non-zeros groups with  small magnitudes. 
However, the ASR of the C\&W attack is quite sensitive to these small
perturbations. As  applying a threshold  to have the same $\ell_0$ norm as our attack, we find that only $6.7\%$ of adversarial examples generated from C\&W attack remain valid.
By contrast, StrAttack
is able to
{highlight the most important group structures (local regions) of adversarial perturbations without attacking other pixels.}
For example, 
 StrAttack
 misclassifies a natural image ($4$ in MNIST) as an incorrect label $3$. That is because the pixels that appears in the   structure of   $3$ are more significantly perturbed   by our attack; see  
 the top right plots of Fig.\,\ref{fig: samples}.  Furthermore, the `goose-sorrel' example shows that misclassification occurs when we just perturb a small number of non-sparse group regions on goose’s head, which is more consistent with human perception.
We refer readers to  Appendix\,\ref{Results} for more results.
 

By quatitatively analysis, we  report $\ell_p$ norms and ASR in Table \ref{table_l2_all} for $p \in \{0,1,2,\infty \}$. We show that 
StrAttack
perturbs much fewer pixels (smaller $\ell_0$ norm), but it
is comparable to or even better than other attacks in terms of $\ell_1$, $\ell_2$, and $\ell_\infty$ norms.
Specifically, the FGM attack yields the worst performance in both ASR and $\ell_p$ distortion. 
On MNIST and CIFAR-10, 
StrAttack
outperforms other attacks in $\ell_0$, $\ell_1$ and $\ell_\infty$ distortion. On ImageNet, 
StrAttack
outperforms C\&W attack in $\ell_0$ and $\ell_1$ distortion.
Since the C\&W attacking loss directly penalizes the $\ell_2$ norm, it often causes smaller  $\ell_2$ distortion than StrAttack. 
{We also observe that the overlapping case leads to the adversarial perturbation of less sparsity (in terms of $\ell_0$ norm) compared to the non-overlapping case. This is not surprising, since the sparsity of the overlapping region is controlled by at least two groups.  However, compared to C\&W attack, the use of overlapping groups in StrAttack still yields sparser perturbations.  Unless specified otherwise, we focus on the case of non-overlapping groups to generate the most sparse adversarial perturbations. 
We highlight that although a so-called one-pixel attack  \citep{su2017one}
  also  yields very small $\ell_0$ norm, it is   
at the cost of very large $\ell_\infty$ distortion. Unlike one-pixel attack, StrAttack achieves the sparsity   \textit{without losing}  the performance of  $\ell_\infty$, $\ell_1$ and $\ell_2$ distortion.}

\textcolor{black}{Furthermore, 
 we compare the performance of StrAttack with the C\&W $\ell_\infty$ attack and IFGSM while attacking the robust model \citep{madry2017towards} on MNIST. We remark that all the considered attack methods are performed under the same  $\ell_\infty$-norm based distortion constraint with an upper bound $\epsilon \in \{ 0.1,0.2,0.3,0.4\}$.
Here we obtain a (refined) StrAttack subject to $\| \boldsymbol{\delta }\|_\infty \leq \epsilon$  by solving problem \eqref{eq: prob_retrain} at $\gamma = 0$. In Table\,\ref{table:attack_madry},  we demonstrate the ASR and the number of perturbed pixels for various attacks  over $5000$ (untargeted) adversarial examples.
The ASR define as the proportion of the final perturbation results less than given $\epsilon \in \{ 0.1,0.2,0.3,0.4\}$ bound over number of test images. Here an successful attack is defined by an attack that can fool DNNs and meets the $\ell_\infty$ distortion constraint.
As we can see, StrAttack can achieve the similar ASR compared to other attack methods, however, it perturbs a much less number of pixels.
\textcolor{black}{Next, we evaluate the performance of StrAttack
against two defense mechanisms: defensive distillation  \citep{papernot2016distillation} and adversarial training \citep{tram2018ensemble}. We observe that StrAttack is able to break the two defense methods with 100\% ASR. More details are provided in Appendix\,\ref{two_defense}.}
}

\textcolor{black}{Lastly, we evaluate the transferability of StrAttack from Inception V3 \citep{Szegedy2016RethinkingTI} to other network models including Inception V2, Inception V4 \citep{szegedy2017inception}, ResNet 50, ResNet 152 \citep{he2016deep}, DenseNet 121 and DenseNet 161 \citep{huang2017densely}.
For comparison, we also present the transferbility of IFGSM and C\&W. This experiment is performed 
under $1000$ (target) adversarial examples on ImageNet\footnote{We follow the experiment setting in \citep{su2018robustness}, where  the transferability is evaluated by the target class top-5 success rate at each transferred model.}. It can be seen from in Table\, \ref{table:attack_transfer} that StrAttack yields the largest attack success rate while transferring to almost every network model.
}






\begin{table*}\small
 \centering
  \caption{Adversarial attack success rate (ASR) and $\ell_p$ distortion values for various attacks. 
  } 
  \label{table_l2_all}
   \vspace*{-0.1in}
\scalebox{0.65}[0.65]{
\begin{threeparttable}
 \begin{tabular}{c|c|ccccc|ccccc|ccccc}
\toprule[1pt]
\multirow {2}{*}{Data Set}  &  \multirow {2}{*}{\makecell{Attack \\ Method}}  & \multicolumn{5}{c|}{\makecell{Best Case\tnote{*}}}   &  \multicolumn{5}{c|}{Average Case\tnote{*}}   &  \multicolumn{5}{c}{Worst Case\tnote{*}}   \\
  \cline{3-17}  
 &  & \multicolumn{1}{c}{ASR} & \multicolumn{1}{c}{${\ell_0}$} & \multicolumn{1}{c}{${\ell_1}$} & \multicolumn{1}{c}{$\ell_2$} & \multicolumn{1}{c|}{$\ell_\infty$} 
  & \multicolumn{1}{c}{ASR} & \multicolumn{1}{c}{${\ell_0}$} & \multicolumn{1}{c}{${\ell_1}$} & \multicolumn{1}{c}{$\ell_2$} & \multicolumn{1}{c|}{$\ell_\infty$} 
  & \multicolumn{1}{c}{ASR}  & \multicolumn{1}{c}{${\ell_0}$} & \multicolumn{1}{c}{${\ell_1}$} & \multicolumn{1}{c}{$\ell_2$} & \multicolumn{1}{c}{$\ell_\infty$}  \\
\midrule[1pt]
\multirow {4}{*}{MNIST}  &  FGM & 99.3 & 456.5  &28.2 &2.32 &0.57 & 35.8 & 466& 39.4 & 3.17 & 0.717 & 0 & N.A.\tnote{**} & N.A. &N.A. & N.A. \\
&  IFGSM & 100 & 549.5 &18.3 &1.57 & 0.4& 100& 588& 30.9&2.41 & 0.566& 99.8& 640.4 & 50.98 & 3.742 & 0.784  \\
 & C\&W & 100 &479.8 & 13.3 & 1.35 & 0.397 & 100 & 493.4 & 21.3  & \textbf{1.9} & 0.528 & 99.7 & 524.3 & 29.9 & \textbf{2.45} & 0.664 \\
&  StrAttack & 100 & \textbf{73.2} & 10.9 & 1.51 & \textbf{0.384} & 100 & \textbf{119.4} & 18.05 & 2.16 & \textbf{0.47}  & 100 & \textbf{182.0} &  26.9  & 2.81 & \textbf{0.5}   \\
 & +overlap & 100 &84.4 & \textbf{9.2} & \textbf{1.32} & 0.401 & 100 & 157.4 & \textbf{16.2}  & 1.95 & 0.508 & 100 & 260.9 & \textbf{22.9} & 2.501 & 0.653 \\
\midrule[1pt]
\multirow {4}{*}{CIFAR-10}&  FGM& 98.5 & 3049 &  12.9 &0.389  & 0.046 & 44.1 &  3048 & 34.2 & 0.989  & 0.113 & 0.2 & 3071 & 61.3 & 1.76 & 0.194 \\
 & IFGSM & 100  &  3051 & 6.22 & 0.182  & 0.02 & 100 & 3051 & 13.7  & 0.391 & 0.0433 & 100 &  3060 & 22.9  & 0.655 & 0.075\\
&  C\&W  & 100  &2954  & 6.03 &0.178   & \textbf{0.019} & 100 & 2956 & 12.1 & 0.347  & \textbf{0.0364} & 99.9 &3070 & 16.8  & \textbf{0.481} & \textbf{0.0536} \\
& StrAttack& 100 & \textbf{264} & \textbf{3.33} & 0.204 & 0.031 &100 & \textbf{487} & 7.13 & 0.353 & 0.050 &100 & \textbf{772} & \textbf{12.5} &0.563 & 0.075 \\ 
 & +overlap & 100 &295 & 3.35 & \textbf{0.169} & 0.029 & 100 & 562 & \textbf{7.05}  & \textbf{0.328} & 0.047 & 100 & 920 & 12.9 & 0.502 & 0.063 \\
\midrule[1pt]
\multirow {4}{*}{ImageNet}& FGM & 12 & 264917 & 152 & 0.477 & \textbf{0.0157}  & 2  & 263585  & 51.3 & \textbf{0.18} & \textbf{0.00614} & 0  & N.A.  & N.A.&  N.A. & N.A. \\
&  IFGSM & 100 & 267079 & 299.32 & 0.9086 & 0.02964 & 100 & 267293 & 723 & 2.2 & 0.0792  & 98 & 267581 & 1378 & 4.22 & 0.158  \\
&  C\&W & 100 &267916 & 127 & \textbf{0.471} & 0.016 &  100 &263140 &198  & 0.679 & 0.03 & 100 &265212 &268   & \textbf{0.852} & \textbf{0.041}\\
& StrAttack &100 & \textbf{14462} & \textbf{55.2} &0.719 &0.058 &100 & \textbf{52328} & \textbf{152} & 1.06& 0.075 & 100& \textbf{80722} & \textbf{197} & 1.35 & 0.122\\
\bottomrule[1pt]
\end{tabular}
\begin{tablenotes}
    \small
    \item[*] \textcolor{black}{Please refer to Appendix\,\ref{parameter_setting} for the definition of best case, best case and worst case. }\\
    \item[**] N.A. means not available in the case of zero ASR, +overlap means structured attack with overlapping groups.
\end{tablenotes}
\end{threeparttable}
}
\vspace{-0.4cm}
\end{table*}

\begin{table}[htb]
\centering
\caption{\textcolor{black}{Attack success rate (ASR) and $\ell_0$ norm of adversarial perturbations for various attacks against robust adversarial training based defense on MNIST.}
}
\label{table:attack_madry}
\begin{adjustbox}{width=0.9\textwidth }
\begin{threeparttable}
\begin{tabular}{ccccc|c}
\toprule[1pt]
\multicolumn{1}{c|}{}   
& ASR at $\epsilon = 0.1$  &  ASR at $\epsilon = 0.2$  &  ASR at $\epsilon = 0.3$  &  ASR at $\epsilon = 0.4$ & $\ell_0$ \\\hline
\multicolumn{1}{c|}{IFGSM }      & 0.01 & 0.02 & 0.09 & 0.94 & 654 \\
\multicolumn{1}{c|}{C\&W $\ell_\infty$ attack}      & 0.01 & 0.02 & 0.10 & 0.96 & 723 \\
\multicolumn{1}{c|}{StrAttack} & 0.01 & 0.02 & 0.10 & 0.99 & 279\\
\bottomrule[1pt]
\end{tabular}
\end{threeparttable}
\end{adjustbox}
\end{table}

\begin{table}[htb]
\centering
\caption{\textcolor{black}{Comparison of transferability of different attacks over 6 ImageNet models. 
}
}
\label{table:attack_transfer}
\begin{adjustbox}{width=0.9\textwidth }
\begin{threeparttable}
\begin{tabular}{ccccccc}
\toprule[1pt]
          & Incept V2 & Incept V4 & ResNet50 & ResNet152 & DenseNet121 & DenseNet161 \\\hline
IFGSM     & 0.27         & 0.22         & \textbf{0.27}         & 0.19          & 0.16        & 0.19        \\
C\&W      & 0.25         & 0.24         & 0.23         & 0.23          & 0.15        & 0.15        \\
StrAttack & \textbf{0.28}         & \textbf{0.27}         & 0.25         & \textbf{0.25}          & \textbf{0.26}        & \textbf{0.25}       \\
\bottomrule[1pt]
\end{tabular}
\end{threeparttable}
\end{adjustbox}
\end{table}


\section{StrAttack Offers Better Interpretability}
\vspace*{-0.1in}
\textcolor{black}{In this section, we 
evaluate the effects of structured adversarial perturbations on image classification  through  adversarial  saliency map  (ASM) \citep{papernot2016limitations} and class activation map (CAM) \citep{zhou2016learning}.  Here we recall that ASM measures the impact of
  pixel-level perturbations on label classification, and CAM  localizes class-specific   image discriminative regions that we use to visually explain adversarial perturbations \citep{DBLP:journals/corr/abs-1801-02612}. We will show that compared to C\&W attack, StrAttack meets better interpretability in terms of (a) a higher ASM score  and (b) a tighter connection with CAM, where the metric (a) implies  interpretability at a micro-level, namely, perturbing  pixels with largest impact on image classification, and the metric (b) demonstrates interpretability at a macro-level, namely, perturbations can be mapped to the most   discriminative image regions localized by CAM.
}

\begin{figure}[htb]  
\vspace*{-0.1in}
  \centering
\includegraphics[width=0.9\textwidth]{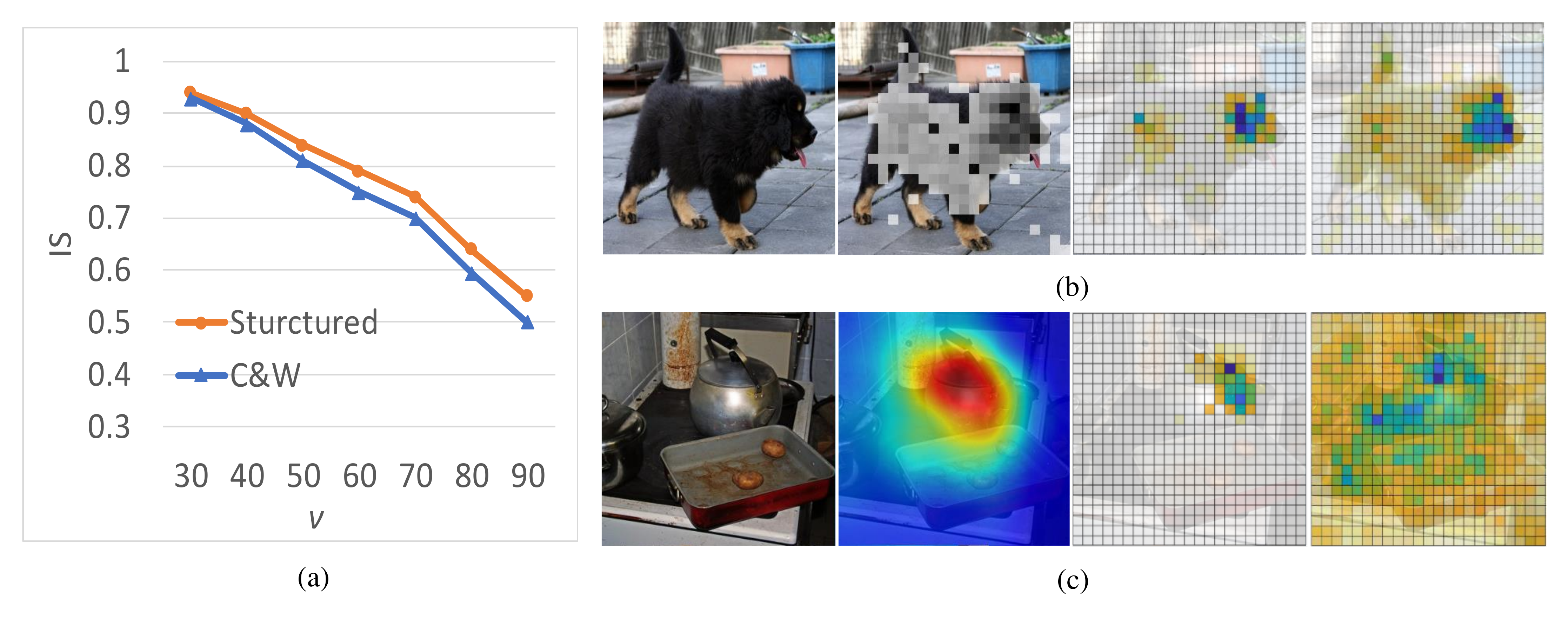}
\vspace*{-0.15in}
\caption{ 
Interpretabilicy comparison of   StrAttack and C\&W attack.  
(a) ASM-based IS vs $\nu$, given  from the $30$th percentile to the $90$th percentile of ASM scores. (b) Overlay ASM and $\mathbf B_{\mathrm{ASM}} \circ  \boldsymbol{\delta}$ on top of image with the true label `Tibetan Mastiff' and the target label `streetcar'. 
From left to right: original image, ASM (darker color represents larger value of ASM score),  $\mathbf B_{\mathrm{ASM}} \circ  \boldsymbol{\delta}$ under StrAttack, and $\mathbf B_{\mathrm{ASM}} \circ  \boldsymbol{\delta}$ under C\&W attack.
Here $\nu$ in $\mathbf B_{\mathrm{ASM}}$ is set by the $90$th percentile of ASM scores. 
(c) From left to right: original image with  true label `stove', 
CAM of `stove',  and  perturbations with target label `water ouzel' under  StrAttack and C\&W.
} 
\label{fig: asmcam}
\end{figure}

\textcolor{black}{Given an input image $\mathbf x_0$ and a target class $t$, let $\mathrm{ASM}(\mathbf x_0, t) \in \mathbb R^d$ denote  ASM scores for every pixel of  $\mathbf x_0$ corresponding to $t$. We
elaborate on the mathematical definition of ASM in Appendix\,\ref{sec: ASM}. Generally speaking, the $i$th element of $\mathrm{ASM}(\mathbf x_0, t)$, denoted by
$\mathrm{ASM}(\mathbf x_0, t)[i]$,
measures how much the classification score with  respect  to the target label $t$  will increase and that
 with  respect  to the original label $t_0$  will decrease  
if  a   perturbation is added to the pixel $i$.
With the aid of ASM, we then define 
a Boolean map $ \mathbf B_{\mathrm{ASM}} \in \mathbb R^d$ to encode the regions of $\mathbf x_0$ most sensitive  to  targeted adversarial attacks, where
$\mathbf B_{\mathrm{ASM}} (i) = 1$ if $\mathrm{ASM}(\mathbf x_0, t) > \nu$, and $0$ otherwise. Here $\nu$ is a given threshold to highlight the most sensitive pixels. 
we then define the  interpretability score (IS) via ASM,
{\small\begin{align}\label{eq: IS}
    \mathrm{IS}(\boldsymbol{\delta}) = {\|\mathbf B_{\mathrm{ASM}} \circ  \boldsymbol{\delta} \|_2}/{\| \boldsymbol{\delta} \|_2},
\end{align}}%
where $\circ$ is the element-wise product. The rationale behind \eqref{eq: IS} is that      $\mathrm{IS}(\boldsymbol{\delta}) \to 1$ if the sensitive region identified by ASM  perfectly predicts  the locations of adversarial perturbations. By contrast, if $\mathrm{IS}(\boldsymbol{\delta}) \to 0$, then   adversarial perturbations cannot be interpreted by ASM.
In Fig.\,\ref{fig: asmcam}(a), we compare IS of our proposed attack with C\&W attack versus the threshold $\nu$, valued by different percentiles of ASM scores.  We obsreve that our attack outperforms C\&W attack in terms of  IS, since the former is able to extract important local structures of images by penalizing the group sparsity of adversarial perturbations.
It seems that our improvement  is not significant. However, 
StrAttack just perturbs very few pixels to obtain this benefit, leading to perturbations with more semantic structure; see Fig.\,\ref{fig: asmcam}(b) for an illustrative example. 
}

\textcolor{black}{Besides ASM, we  show that  the effect of adversarial perturbations can be visually explained through the class-specific discriminative image regions localized by CAM \citep{zhou2016learning}. In Fig.\,\ref{fig: asmcam}(c), we illustrate CAM and demonstrate the differences between our attack and C\&W   in terms of their connections to the most discriminative regions of $\mathbf x_0$ with label $t_0$. We observe that the mechanism of StrAttack can be better interpreted from CAM: only a few adversarial perturbations are needed to suppress the  feature of the original image with the true label. By replacing ASM with CAM, we can similarly compute IS  in \eqref{eq: IS} averaged over $500$ examples on ImageNet, yielding $0.65$ for C\&W attack and $0.77$ for our attack. More examples of ASM and CAM can be viewed in Appendix\,\ref{sec: ASM}.
}

\begin{figure}
   \centering
\hspace*{-0.15in}\begin{tabular}{p{0.56in}p{0.74in}p{0.74in}p{0.56in}p{0.56in}p{0.56in}p{0.56in}}
  ~~~\makecell{\footnotesize    original }
  & ~~~\makecell{ \footnotesize  C\&W  } 
 & ~~~\makecell{ \footnotesize StrAttack  } 
 & ~~ \makecell{ \footnotesize $C(\bm x_0$,$t_0)$ }
 & ~~\makecell{\footnotesize     $C(\bm x_0$,$t)$ } 
 & \makecell{ \footnotesize     $C(\bm x_{\mathrm{CW}}$,$t)$}  \hspace*{-0.15in}
  & \makecell{ \footnotesize     $C(\bm x_{\mathrm{Str}}$,$t)$ }  \hspace*{-0.15in}
\\
\includegraphics[width=0.7in]{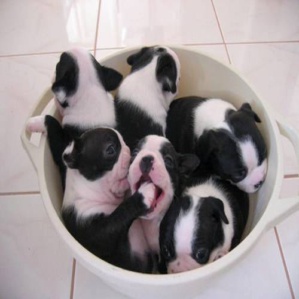}&  
\includegraphics[height = 0.70in, width=0.86in]{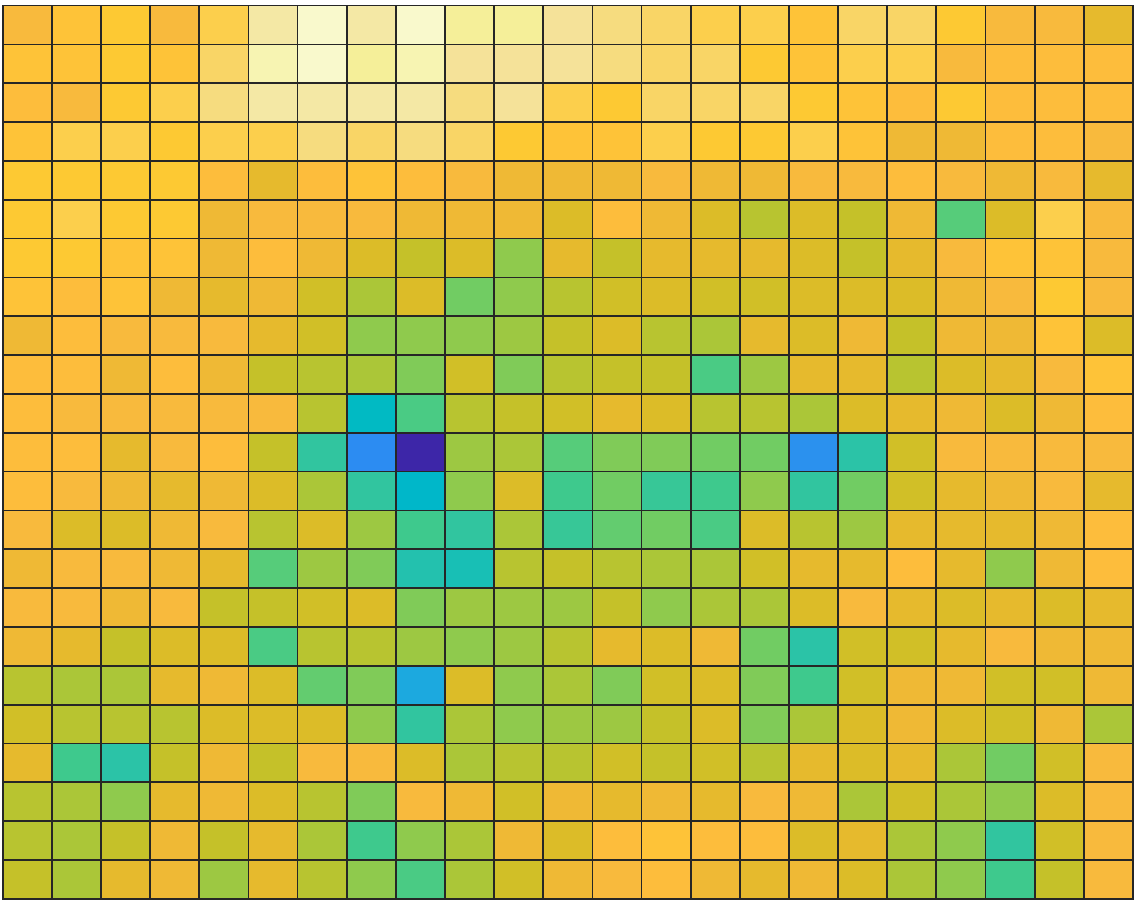}& 
\includegraphics[height = 0.70in,width=0.86in]{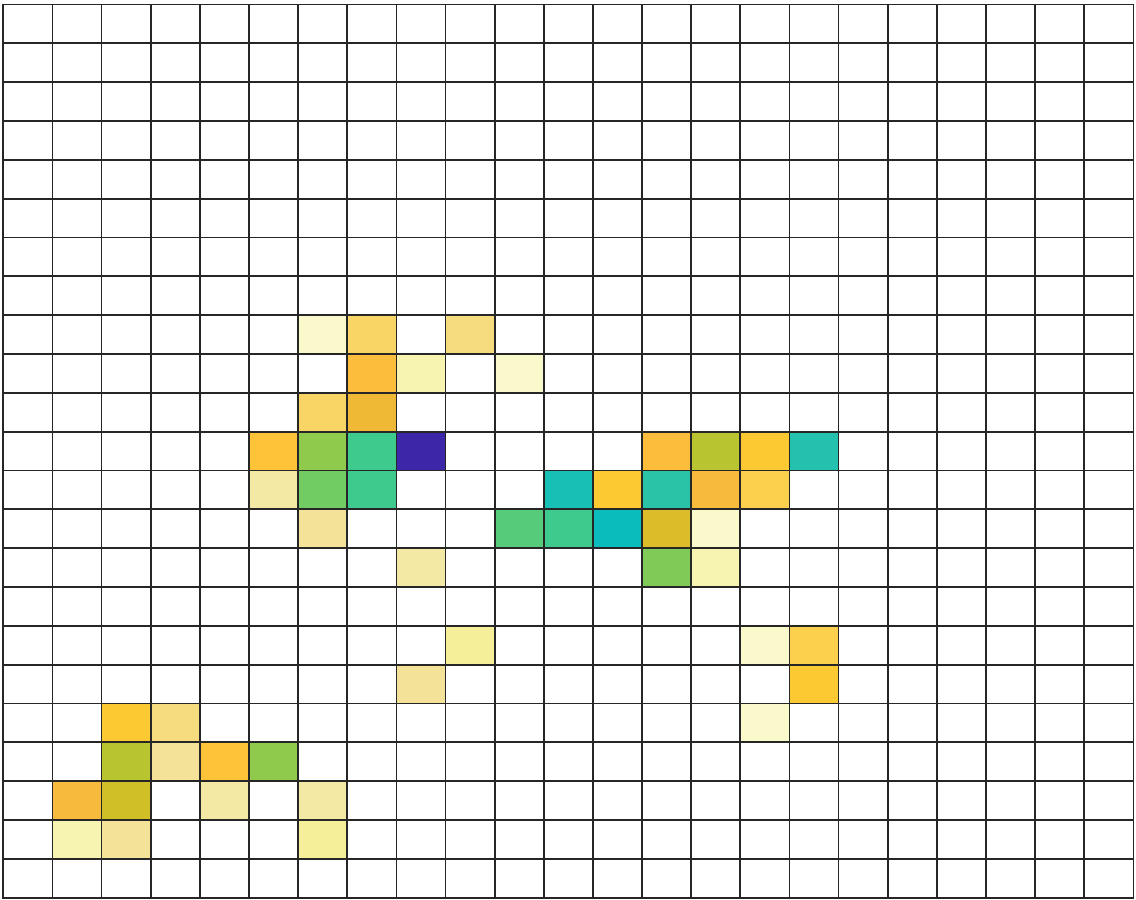}&
\includegraphics[width=0.7in]{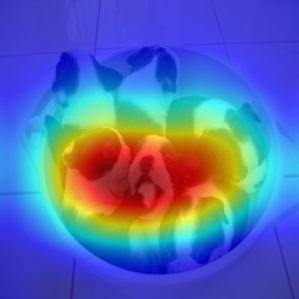}& 
\includegraphics[width=0.7in]{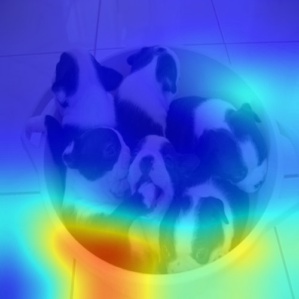}&
\includegraphics[width=0.7in]{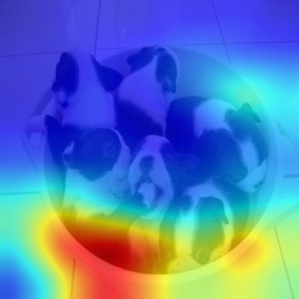}&
\includegraphics[width=0.7in]{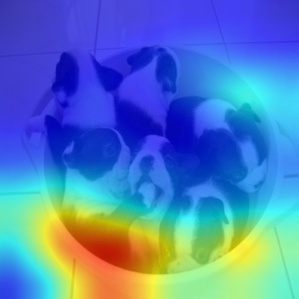} 
\\
\multicolumn{3}{l}{$t_0$: Boston bull, $t$: bucket} & & 
\\[0.3cm]
\includegraphics[width=0.7in]{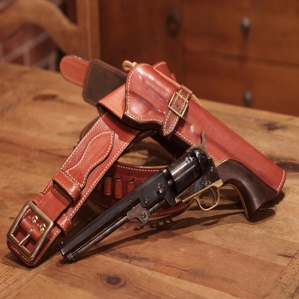}&  
\includegraphics[height = 0.70in, width=0.86in]{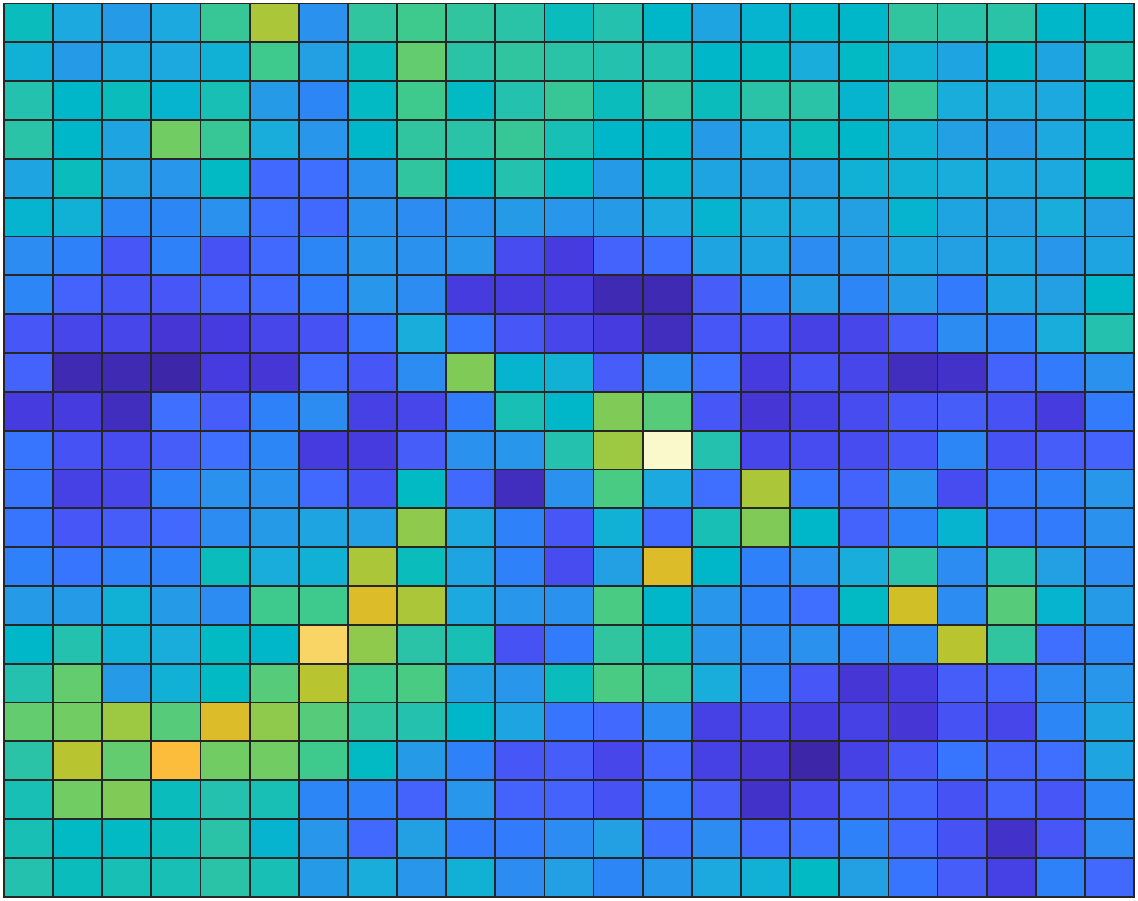}& 
\includegraphics[height = 0.70in, width=0.86in]{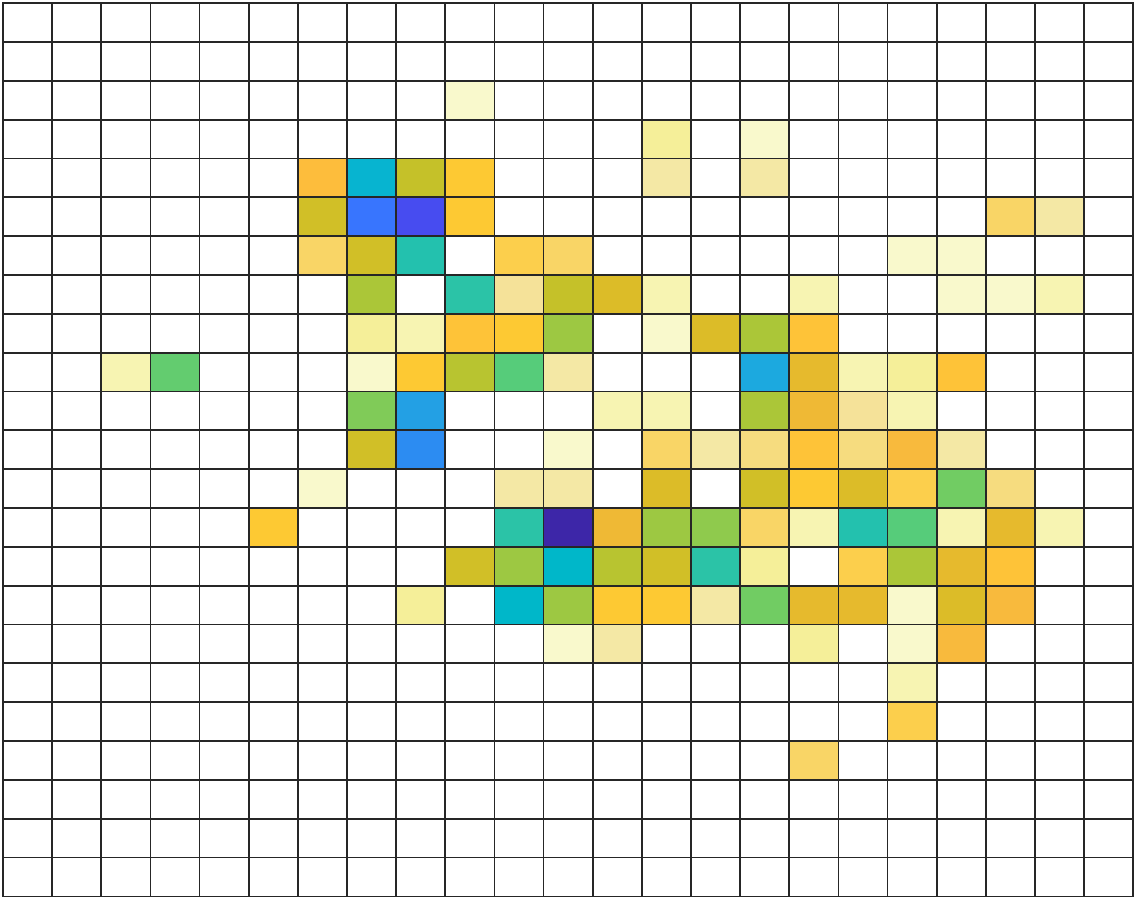}&
\includegraphics[width=0.7in]{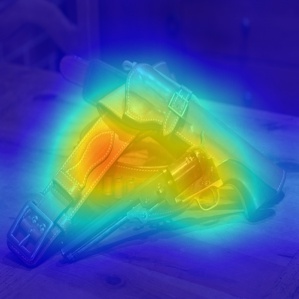}& 
\includegraphics[width=0.7in]{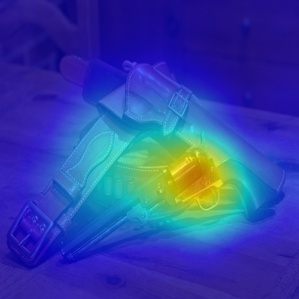}&
\includegraphics[width=0.7in]{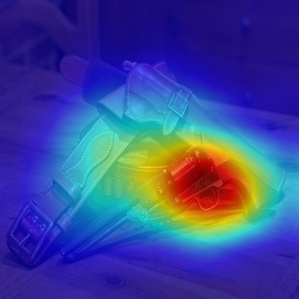}&
\includegraphics[width=0.7in]{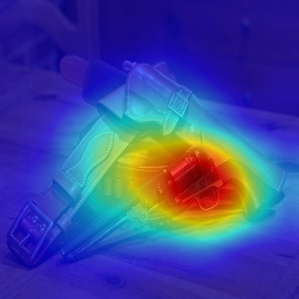} 
\\
\multicolumn{3}{l}{$t_0$: holster, $t$: revolver} & & 
\\[0.3cm]
\includegraphics[width=0.7in]{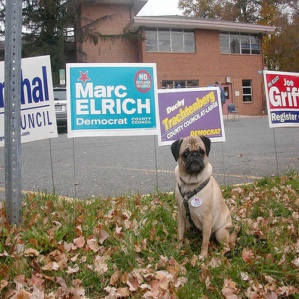}&  
\includegraphics[height = 0.70in, width=0.86in]{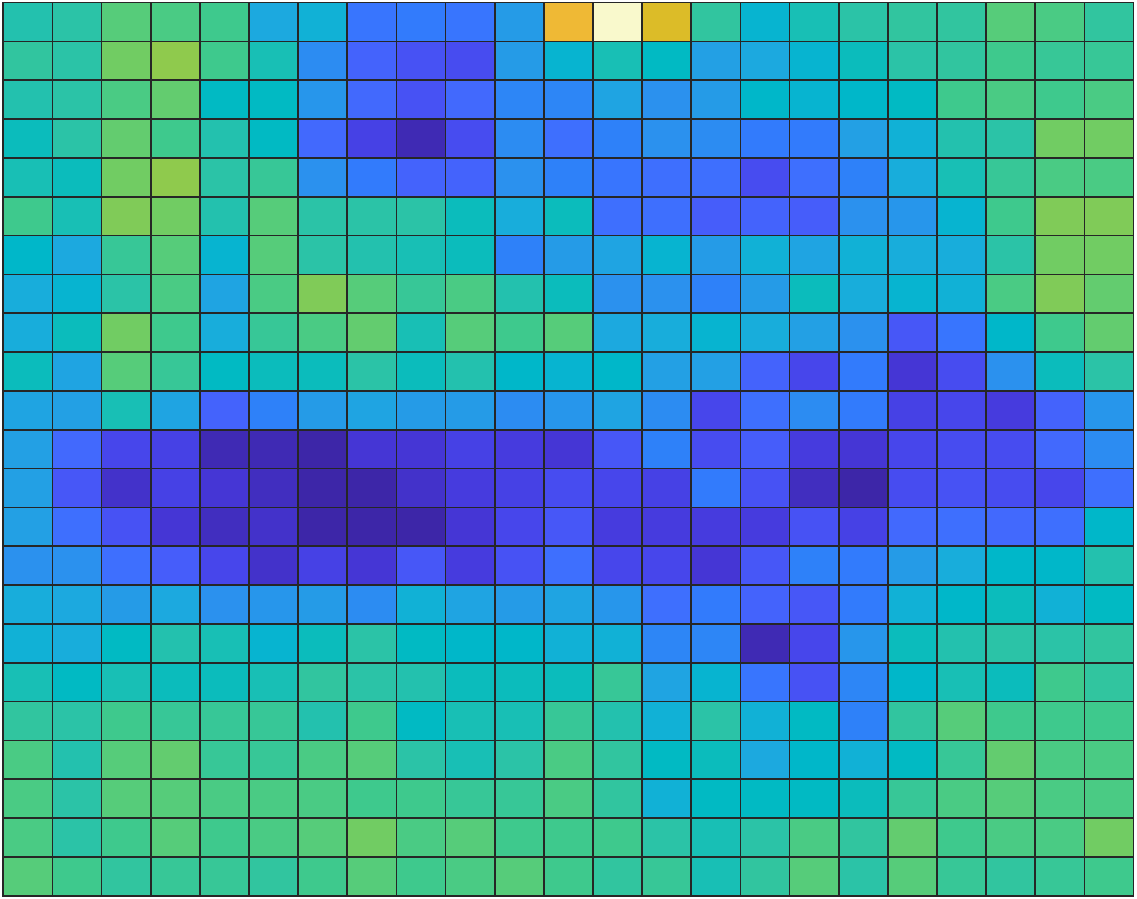}& 
\includegraphics[height = 0.70in, width=0.86in]{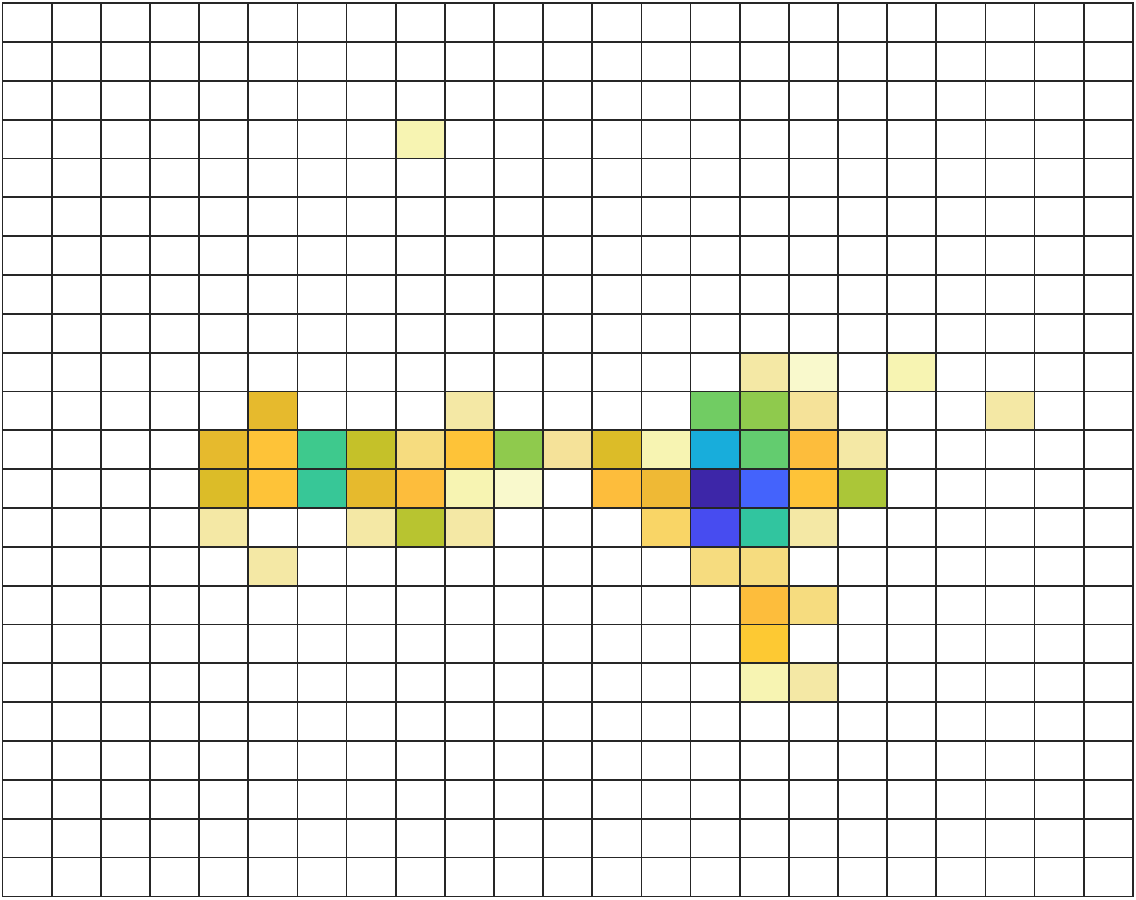}&
\includegraphics[width=0.7in]{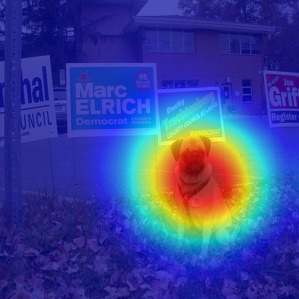}& 
\includegraphics[width=0.7in]{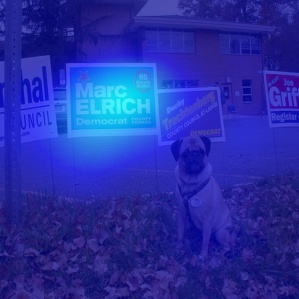}&
\includegraphics[width=0.7in]{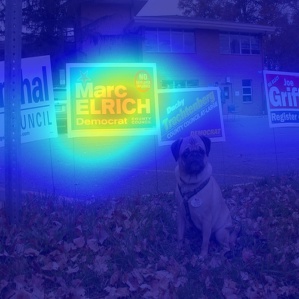}&
\includegraphics[width=0.7in]{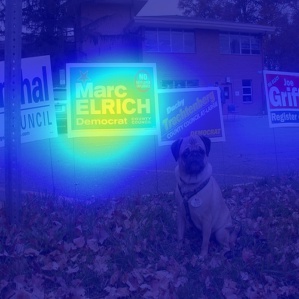} \\
\multicolumn{3}{l}{$t_0$: pug, $t$: street sign} & & 
\vspace*{0.1in} 
\end{tabular}
\caption{\footnotesize{
\textcolor{black}{CAMs of adversarial examples 
generated by the  C\&W attack and   StrAttack under the original label ($t_0$) and   target label ($t$).
Here  $\bm x_{\mathrm{CW}}$ and $\bm x_{\mathrm{Str}}$ denote    adversarial examples crafted by  different attacks. 
At each row, the subplots from left to right represent the original image $\bm x_0$,  perturbations generated by C\&W attack, perturbations generated by StrAttack, and CAMs with respect to  natural or adversarial example under $t_0$ or $t$. Here the class $c$ specified CAM with respect to image $\mathbf x$ is denoted by $C(\mathbf x, c)$.}
}}
    \label{fig: complex_examples}
\end{figure}

\textcolor{black}{
To better interpret the mechanism of adversarial examples, we study adversarial attacks on some  complex images, where the objects of the original and target labels exist simultaneously as shown in Fig.\,\ref{fig: complex_examples}. It can be visualized from CAM that 
both C\&W attack and StrAttack yields similar adversarial effects on natural images: Adversarial perturbations are used to suppress the most discriminative region with respect to the true label, and simultaneously promotes the discriminative region of the target label. The former principle is implied by the location of perturbed regions and $C(\mathbf x_0, t_0)$ in Fig.\,\ref{fig: complex_examples}, and the latter  can be seen from $C(\mathbf x_{\mathrm{CW}}, t)$ or $C(\mathbf x_{\mathrm{Str}}, t)$ against $C(\mathbf x_0, t)$. 
However, compared to C\&W attack, StrAttack perturbs much less but `right' pixels which have 
better correspondence with class-specific discriminative image regions localized by CAM.
}

\vspace*{-0.1in}
\section{Conclusion}
\vspace*{-0.1in}
This work explores group-wise sparse structures when implementing adversarial attacks.
Different from previous works that use $\ell_p$ norm to measure the similarity between an original image and an adversarial example, this work incorporates group-sparsity regularization into the problem formulation of generating adversarial examples and achieves strong group sparsity in the obtained adversarial perturbations.
Leveraging ADMM, we  develop an efficient implementation to generate structured adversarial perturbations, which can be further used to refine an arbitrary adversarial attack under fixed group sparse structures.
The proposed ADMM framewrok is general enough for implementing many state-of-the-art attacks.
We perform extensive experiments using MNIST, CIFAR-10 and ImageNet datasets, showing that our structured adversarial attack (StrAttack) is much stronger than the existing attacks and its better interpretability from group sparse structures aids in uncovering the origins of adversarial examples.

\section*{Acknowledgement}
This work is supported by Air Force Research Laboratory FA8750-18-2-0058, and U.S. Office of Naval Research.

\bibliographystyle{iclr2019_conference}
\bibliography{iclr2019_conference}

\newpage
\appendix
\setcounter{section}{0}
\setcounter{figure}{0}
\makeatletter 
\renewcommand{\thefigure}{A\@arabic\c@figure}
\makeatother
\setcounter{table}{0}
\renewcommand{\thetable}{A\arabic{table}}

\section*{Appendix}

\section{Illustrative Example of Group Sparsity}

 \begin{figure}[htb]  
  \centering
\includegraphics[width=.8\textwidth]{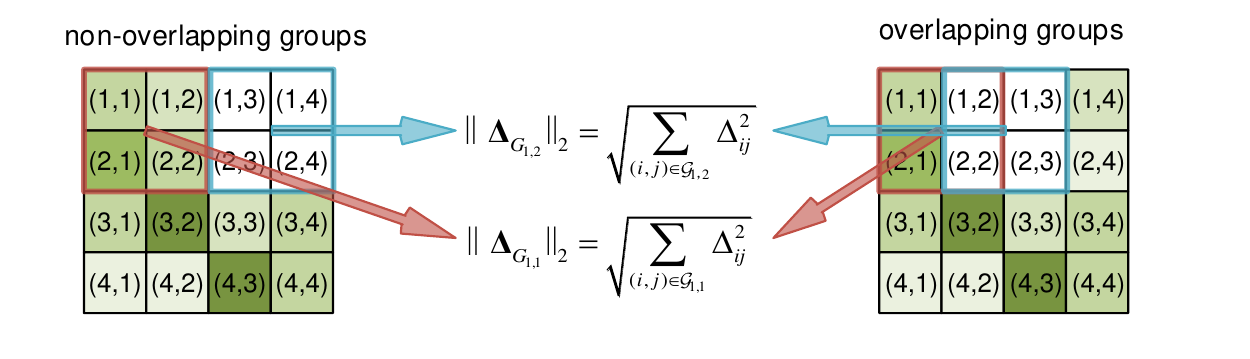}
\caption {An example of $4 \times 4$ perturbation matrix under sliding masks with different strides. The values of matrix elements are represented by color's intensity (white stands for $0$). Left: Non-overlapping groups with $r = 2$ and $S = 2$. Right: Overlapping groups with $r = 2$ and $S = 1$. In both cases, two groups $\mathcal G_{1,1}$ and $\mathcal G_{1,2}$ are highlighted, where $\mathcal G_{1,1}$ is non-sparse, and $\mathcal G_{1,2}$ is sparse.
}
\label{fig: overlap}
\end{figure}

\section{Proof of Proposition\,\ref{prop: prop1}} \label{proof_Pro_1}
We recall that the augmented Lagrangian function 
$L(\boldsymbol \delta, \mathbf z, \mathbf w, \mathbf y, \mathbf u, \mathbf v, \mathbf s)$  is given by
\begin{align}\label{eq: aug_Lag_non_2}
    L( \mathbf z, \boldsymbol \delta,  \mathbf y, \mathbf w, \mathbf u, \mathbf v, \mathbf s) =&  f( \mathbf z + \mathbf x_0 ) + \gamma
   D(\boldsymbol \delta) + \tau  \textstyle \sum_{i=1}^{PQ} \| \mathbf y_{\mathcal D_i}\|_2 + h(\mathbf w) + \mathbf u^T (\boldsymbol \delta - \mathbf z) \nonumber \\
   & \hspace*{-0.5in} + \mathbf v^T (\mathbf y -  \mathbf z)+ \mathbf s^T (\mathbf w - \mathbf z)+ \frac{\rho}{2} \| \boldsymbol \delta - \mathbf z\|_2^2 + \frac{\rho}{2} \| \mathbf y -  \mathbf z \|_2^2 + \frac{\rho}{2} \| \mathbf w - \mathbf z\|_2^2.
\end{align}
Problem \eqref{eq: delta_w_step}, to minimize $ L(\boldsymbol \delta, \mathbf z^k, \mathbf w, \mathbf y, \mathbf u^k, \mathbf v^k, \mathbf s^k)$, can be decomposed into   three sub-problems: 
\begin{align}\label{eq: delta_D}
    \displaystyle \minimize_{\boldsymbol \delta} ~ \gamma D(\boldsymbol{\delta}) + \frac{\rho}{2} \| \boldsymbol{\delta} - \mathbf a \|_2^2,
\end{align}
\begin{align}\label{eq: w_prob}
    \displaystyle \minimize_{\mathbf w} ~ h(\mathbf w) + \frac{\rho}{2} \| \mathbf w - \mathbf b \|_2^2,
\end{align}
\begin{align}\label{eq: y_prob}
    \displaystyle \minimize_{\mathbf y} ~ \tau   \displaystyle \sum_{i=1}^{PQ} \| \mathbf y_{\mathcal D_i}\|_2 + \frac{\rho}{2} \| \mathbf y -  \mathbf c \|_2^2,
\end{align}
where $\mathbf a \Def \mathbf z^k -  \mathbf u^k/\rho$, $\mathbf b \Def \mathbf z^k  -  \mathbf s^k / \rho $, and  $\mathbf c \Def  \mathbf z^k -   \mathbf v^k / \rho $.
 
\paragraph{$\boldsymbol \delta$-step} 
Suppose $D(\boldsymbol \delta) = \| \boldsymbol \delta \|_2^2$, then the solution to problem \eqref{eq: delta_D} is easily acquired as below 
\begin{align} \label{delta_solution}
    \boldsymbol \delta^{k+1} = \frac{\rho}{\rho + 2\gamma} \mathbf a 
\end{align}
 


\paragraph{$\mathbf w$-step}
Based on the definition of $h(\mathbf w)$, problem \eqref{eq: w_prob} becomes
\begin{align}\label{eq: w_prob_sm}
\begin{array}{ll}
       \displaystyle \minimize_{\mathbf w} &  \| \mathbf w - \mathbf b \|_2^2 \\
     \st & (\mathbf x_0 + \mathbf w) \in [0,1]^n, ~ \| \mathbf w \|_\infty \leq \epsilon.
\end{array}
\end{align}
Problem \eqref{eq: w_prob_sm} is equivalent to
\begin{align}\label{eq: w_prob_sm_sig}
\begin{array}{ll}
       \displaystyle \minimize_{w_i} &  ( w_i - a_i )_2^2 \\
     \st & - [\mathbf x_0]_i \leq w_i \leq 1-[\mathbf x_0]_i, ~ |w_i| \leq \epsilon
\end{array}
\end{align}
for $i \in [n]$, 
where $x_i$ or $[\mathbf x]_i$ represents the $i$th element of $\mathbf x$, and $ 1-[\mathbf x_0]_i > 0$ since $[\mathbf x_0]_i \in [0,1]$. Problem \eqref{eq: w_prob_sm_sig} then 
yields the solution 
\begin{align} 
    [\mathbf w^{k+1}]_i = \left \{
    \begin{array}{ll}
       \min \{ 1-[\mathbf x_0]_i, \epsilon \}  &  a_i >  \min \{ 1-[\mathbf x_0]_i, \epsilon \} \\
     \max \{ - [\mathbf x_0]_i, -\epsilon \}   &  a_i <  \max \{ - [\mathbf x_0]_i, -\epsilon \}  \\
     a_i & \text{otherwise}.
    \end{array}
    \right.
    \label{w_solution}
\end{align}

\paragraph{$\mathbf y$-step}

Problem \eqref{eq: y_prob} becomes
\begin{align}
\begin{array}{cc}
\displaystyle \minimize_{\mathbf y} & \displaystyle \sum_{i=1}^{PQ} \| \mathbf y_{\mathcal D_i} \|_2 + \frac{\rho}{2\tau} \| \mathbf y - \mathbf c\|_2^2, 
\end{array}
\end{align}
The solution is given by the proximal operator   associated with the    $\ell_2$ norm with parameter $\tau / \rho$ \citep{parikh2014proximal}
\begin{align} 
[ \mathbf y^{k+1} ]_{\mathcal D_i} = \left (
1 - \frac{\tau}{\rho \|  [ \mathbf c ]_{\mathcal D_i} \|_2}
\right )_+ [  \mathbf c  ]_{\mathcal D_i},~  i \in [PQ],
\end{align}
where recall that $\cup_{i \in [PQ]} \mathcal D_i = [n]$, and $\mathcal D_i \cap \mathcal D_j = \emptyset$ if $i \neq j$.
\hfill $\square$



\section{Proof of Proposition\,\ref{prop: prop2}} \label{proof_Pro_2}
The augmented Lagrangian of problem \eqref{eq: prob_ADMM_overlap} is given by
\begin{align}
    L(\mathbf z, \boldsymbol \delta,  \mathbf w, \{ \mathbf y_i \}, \mathbf u, \mathbf v_i, \mathbf s) =&  f( \mathbf z + \mathbf x_0 ) + \gamma
   D(\boldsymbol \delta)+ h(\mathbf w)  + \tau \sum_{i=1}^{PQ} \| \mathbf y_{i,\mathcal D_i}\|_2  + \mathbf u^T (\boldsymbol \delta - \mathbf z) + \mathbf s^T (\mathbf w - \mathbf z)  \nonumber \\
   &+ \sum_{i=1}^{PQ} \mathbf v_i^T (\mathbf y_i - \mathbf z)  + \frac{\rho}{2} \| \boldsymbol \delta - \mathbf z\|_2^2  + \frac{\rho}{2} \| \mathbf w - \mathbf z\|_2^2 + \frac{\rho}{2}  \sum_{i=1}^{PQ} \| \mathbf y_i - \mathbf z \|_2^2,
\end{align}
where $\mathbf u$, $\mathbf v_i$ and $\mathbf s$ are the Lagrangian multipliers.

ADMM decomposes the optimization variables into \textit{two} blocks and  adopts  the following iterative scheme,
\begin{align}
   & \{ \boldsymbol \delta^{k+1}, \mathbf w^{k+1}, \mathbf y_i^{k+1} \} = \argmin_{\boldsymbol{\delta},\mathbf w, \{ \mathbf y_i \} } L( \mathbf z^k, \boldsymbol \delta, \mathbf w, \mathbf y_i, \mathbf u^k, \mathbf v_i^k, \mathbf s^k), \label{eq: delta_w_step_overlap}\\
   &   \mathbf z^{k+1}= \argmin_{\mathbf z} L( \mathbf z, \boldsymbol \delta^{k+1}, \mathbf w^{k+1}, \mathbf y_i^{k+1}, \mathbf u^k, \mathbf v_i^k, \mathbf s^k),  \label{eq: z_step_overlap}\\
   & \left \{
   \begin{array}{l}
        \mathbf u^{k+1} = \mathbf u^k + \rho (\boldsymbol \delta^{k+1} - \mathbf z^{k+1}), \\
         \mathbf v_i^{k+1} = \mathbf v_i^k + \rho (\mathbf y_i^{k+1} - \mathbf z^{k+1}),  ~\text{for}~ i \in [PQ], \\
        \mathbf s^{k+1} = \mathbf s^k + \rho (\mathbf w^{k+1} - \mathbf z^{k+1}) ,
   \end{array}
   \right.
   \label{eq: dual_update_overlap}
\end{align}
where $k$ is the iteration index. Problem \eqref{eq: delta_w_step_overlap} can be split into three subproblems as shown below,
\begin{align}\label{eq: delta_D_overlap}
    \displaystyle \minimize_{\boldsymbol \delta} ~ \gamma D(\boldsymbol{\delta}) + \frac{\rho}{2} \| \boldsymbol{\delta} - \mathbf a \|_2^2,
\end{align}
\begin{align}\label{eq: w_prob_overlap}
    \displaystyle \minimize_{\mathbf w} ~ h(\mathbf w) + \frac{\rho}{2} \| \mathbf w - \mathbf b \|_2^2,
\end{align}
\begin{align}\label{eq: y_prob_overlap}
    \displaystyle \minimize_{\mathbf y_i} ~ \tau  \displaystyle  \| \mathbf y_{i,\mathcal D_i}\|_2 + \frac{\rho}{2}  \| \mathbf y_i - \mathbf c_i\|_2^2, ~\text{for}~ i \in [PQ].
\end{align}
where $\mathbf a = \mathbf z^k - \mathbf u^k/\rho$, $\mathbf b = \mathbf z^k - \mathbf s^k/\rho$ and $\mathbf c_i =  \mathbf z^k  - \mathbf v_i^k/\rho $.
Each problem has a closed form solution. Note that the solutions to problem \eqref{eq: delta_D_overlap} and problem \eqref{eq: w_prob_overlap}   are given \eqref{delta_solution} and \eqref{w_solution}. 


\paragraph{$\mathbf y_i$-step}
Problem \eqref{eq: y_prob_overlap} can be rewritten as
\begin{align}\label{eq: y_prob_overlap_2}
    \displaystyle \minimize_{\mathbf y_i} ~ \tau  \displaystyle  \| \mathbf y_{i,\mathcal D_i}\|_2 + \frac{\rho}{2}  \| \mathbf y_{i,\mathcal D_i} - [ \mathbf c_i ]_{\mathcal D_i}\|_2^2 + \frac{\rho}{2}  \|  \mathbf y_{i,[n]/\mathcal D_i} - [ \mathbf c_i ]_{[n]/\mathcal D_i}   \|_2^2, ~\text{for}~ i \in [PQ],
\end{align}
which can be 
 decomposed into
\begin{align}\label{eq: y_prob_overlap_21}
    \displaystyle \minimize_{\mathbf y_{i,\mathcal D_i}} ~ \tau  \textstyle  \| \mathbf y_{i,\mathcal D_i}\|_2 + \frac{\rho}{2}  \| \mathbf y_{i,\mathcal D_i} - [ \mathbf c_i ]_{\mathcal D_i}   \|_2^2, ~\text{for}~ i \in [PQ],
\end{align}
and
\begin{align}\label{eq: y_prob_overlap_22}
    \displaystyle \minimize_{\mathbf y_{i,[n]/\mathcal D_i}} ~ \|  \mathbf y_{i,[n]/\mathcal D_i} - [ \mathbf c_i ]_{[n]/\mathcal D_i}   \|_2^2, ~\text{for}~ i \in [PQ].
\end{align}
The solution to problem \eqref{eq: y_prob_overlap_21} can be obtained through the block soft thresholding operator \citep{parikh2014proximal},
\begin{align}\label{eq: y_solution_overlap_21}
 \left [ \mathbf y_i^{k+1} \right  ]_{\mathcal D_i} = \left ( 1 - \frac{\tau}{\rho \|  [ \mathbf c_i ]_{\mathcal D_i} \|_2}
 \right )_+ \left [  \mathbf c_i \right ]_{\mathcal D_i}, ~\text{for}~ i \in [PQ],
\end{align}
The solution to problem \eqref{eq: y_prob_overlap_22} is given by,
\begin{align}\label{eq: y_solution_overlap_22}
  \left [ \mathbf y_i^{k+1} \right ]_{[n]/\mathcal D_i} = \left [ \mathbf c_i \right ]_{[n]/\mathcal D_i}, ~\text{for}~ i \in [PQ].
\end{align}

\paragraph{$\mathbf z$-step}
Problem \eqref{eq: z_step_overlap} can be simplified to 
\begin{align}\label{eq: prob_z_complete_overlap}
    \begin{array}{ll}
\displaystyle \minimize_{\mathbf z}         &  \displaystyle f(\mathbf x_0 + \mathbf z) + \frac{\rho}{2} \|    \mathbf z - \mathbf a^\prime \|_2^2 
+
\frac{\rho}{2}
\|    \mathbf z - \mathbf b^\prime \|_2^2 +   \frac{\rho}{2} \sum_{i=1}^{PQ} \| \mathbf z -  \mathbf c_i^\prime \|_2^2,
    \end{array}
\end{align}
where
$
\mathbf a^\prime \Def \boldsymbol{\delta}^{k+1} +   \mathbf u^k / \rho
$, 
$
\mathbf b^\prime \Def \mathbf w^{k+1}  +   \mathbf s^k / \rho
$,
and $\mathbf c_i^\prime \Def \mathbf y_i^{k+1} +   \mathbf v_i^k/\rho
$.
We solve problem \eqref{eq: prob_z_complete_overlap} using the linearization technique  \citep{suzuki2013dual,liu2017zeroth,boyd2011distributed}.
More specifically, the function $f$ is replaced with its first-order Taylor expansion  at the point $\mathbf z^k$ by adding a Bregman divergence term $(\eta_k/2 )\| \mathbf z - \mathbf z^k \|_2^2$.
As a result, problem \eqref{eq: prob_z_complete_overlap}  becomes 
\begin{align}
    \begin{array}{ll}
\displaystyle \minimize_{\mathbf z}         &  \displaystyle (\nabla f(\mathbf z^k + \mathbf x_0))^T ( \mathbf z  - \mathbf z^k) + \frac{\eta_k}{2} \| \mathbf z  - \mathbf z^k \|_2^2 + \frac{\rho}{2} \|  \mathbf z - \mathbf a^\prime \|_2^2  \\
&\displaystyle +
\frac{\rho}{2}
\|   \mathbf z- \mathbf b^\prime \|_2^2 +   \frac{\rho}{2} \sum_{i=1}^{PQ} \| \mathbf z - \mathbf c_i^\prime \|_2^2,
    \end{array}
    \label{eq: prob_theta_complete_linear_overlap}
\end{align}
whose solution is given by
\begin{align}
\displaystyle
\mathbf z^{k+1} = \frac{ \eta_k \mathbf z^k + \rho \mathbf a^\prime + \rho \mathbf b^\prime +\rho  \sum_{i=1}^{PQ}   \mathbf c_i^\prime- \nabla f(\mathbf z^k + \mathbf x_0) }{\eta_k + (2+PQ) \rho }. \label{eq: z_sol_overlap_appendix}
\end{align}
\hfill $\square$

\section{Proof of Proposition\,\ref{prop: prop3}}
\label{supp: prop3}
We start by converting problem \eqref{eq: prob_retrain} into the ADMM form
 \begin{align} 
\begin{array}{ll}
    \displaystyle \minimize_{ \boldsymbol \delta, \mathbf z} & f(\mathbf x_0 + \mathbf z) + g(\mathbf z) + \gamma D(\boldsymbol \delta)  + h(\boldsymbol{\delta}) + g(\boldsymbol{\delta}) \\
  \st    &   \boldsymbol{\delta} = \mathbf z ,
\end{array}
\end{align}
where $\mathbf z$ and $\boldsymbol{\delta}$ are optimization variables, 
\textcolor{black}{$g(\mathbf \delta)$} is an indicator function with respect to the constraint $\{ \delta_i = 0, ~\text{if}~ i \in \mathcal S_\sigma \}$, and $h(\boldsymbol{\delta})$ is the other indicator function with respect to the other constraints $(\mathbf x_0 + \boldsymbol \delta) \in [0,1]^n, ~ \| \boldsymbol \delta \|_\infty \leq \epsilon$.

The augmented Lagrangian of problem \eqref{eq: prob_retrain} is given by
\begin{align}
    L(\boldsymbol \delta, \mathbf z, \mathbf u) =&   f(\mathbf z + \mathbf x_0 ) + 
    g(\mathbf z) + 
    \gamma D(\boldsymbol \delta) + h(\boldsymbol{\delta}) + g(\boldsymbol{\delta})  + \mathbf u^T (\boldsymbol{\delta} - \mathbf z)+ \frac{\rho}{2} \| \boldsymbol{\delta} - \mathbf z\|_2^2,
\end{align}
where $\mathbf u$ is the Lagrangian multiplier. 
     
ADMM yields the following alternating steps
\begin{align}
   &  \boldsymbol{\delta}^{k+1}  = \argmin_{\boldsymbol{\delta}} L(\boldsymbol{\delta}, \mathbf z^k, \mathbf u^k) \label{eq: delta_w_step_smooth}\\
   &   \mathbf z^{k+1}= \argmin_{\mathbf z} L(\boldsymbol{\delta}^{k+1}, \mathbf z,  \mathbf u^k) \label{eq: z_step_smooth} \\
   &   \mathbf u^{k+1} = \mathbf u^k + \rho (\boldsymbol{\delta}^{k+1} - \mathbf z^{k+1}).
   \label{eq: dual_update_smooth}
\end{align}

\paragraph{$\boldsymbol{\delta}$-step}

Suppose $D(\boldsymbol{\delta}) = \| \boldsymbol{\delta} \|_2^2$,
problem \eqref{eq: delta_w_step_smooth} becomes
\begin{align}\label{eq: delta_D_smooth_v2}
\begin{array}{ll}
    \displaystyle \minimize_{\boldsymbol{\delta}}  & \gamma \| \boldsymbol{\delta} \|_2^2  + \frac{\rho}{2} \| \boldsymbol{\delta}- \mathbf a \|_2^2 \\
    \st    &  (\mathbf x_0 + \boldsymbol{\delta}) \in [0,1]^n,~ \| \boldsymbol{\delta} \|_\infty \leq  \epsilon \\
     & \delta_i = 0, ~\text{if}~ i \in \mathcal S_\sigma,
\end{array}
\end{align}
where $\mathbf a \Def \mathbf z^k - \mathbf u^k/\rho$.
Problem \eqref{eq: delta_D_smooth_v2} can be decomposed elementwise
\begin{align}\label{eq: delta_D_smooth_v2_ele}
\begin{array}{ll}
    \displaystyle \minimize_{\delta_i}  &  \frac{2\gamma+\rho}{\rho}\delta_i^2 - 2 a_i \delta_i + a_i^2 =  \frac{2\gamma+\rho}{\rho} \left (\delta_i - \frac{\rho}{2\gamma+\rho} a_i \right )^2 \\
    \st    &  ( \left [ \mathbf x_0  \right ]_i +  \delta_i ) \in [0,1],  ~ |\delta_i| \leq  \epsilon \\
        & \delta_i = 0, ~\text{if}~ i \in \mathcal S_\sigma.
\end{array}
\end{align}

The solution to problem \eqref{eq: delta_D_smooth_v2_ele} is then given by
\begin{align}
    [\boldsymbol{\delta}^{k+1}]_i = \left \{
    \begin{array}{ll}
     0 & i \in \mathcal S_\sigma \\
       \min \{ 1- \left [\mathbf x_0 \right ]_i, \epsilon \}  &  
       \frac{\rho}{2\gamma+\rho} a_i >  \min \{ 1- \left [ \mathbf x_0  \right ]_i, \epsilon \},~ i \notin \mathcal S_\sigma \\
     \max \{ -  \left [ \mathbf x_0  \right ]_i, -\epsilon \}   &  
     \frac{\rho}{2\gamma+\rho} a_i <  \max \{ -  \left [ \mathbf x_0  \right ]_i, -\epsilon \}, i \notin \mathcal S_\sigma  \\
        \frac{\rho}{2\gamma+\rho} a_i & \text{otherwise}.
    \end{array}
    \right.
\end{align}

\paragraph{$\mathbf z$-step}
Problem \eqref{eq: z_step_smooth} yields
\begin{align}\label{eq: prob_z_smooth_retrain}
    \begin{array}{ll}
\displaystyle \minimize_{\mathbf z}         & \displaystyle  f(\mathbf x_0 + \mathbf z) + \frac{\rho}{2} \|    \mathbf z - \mathbf a^\prime \|_2^2 \\
\st & z_i = 0, ~\text{if}~ i \in \mathcal S_\sigma,
    \end{array}
\end{align}
where $\mathbf a^\prime = \boldsymbol{\delta}^{k+1} + \mathbf u^k/\rho$.
We solve problem \eqref{eq: prob_z_smooth_retrain} using the linearization technique  \citep{suzuki2013dual,liu2017zeroth,boyd2011distributed},
\begin{align}\label{eq: prob_z_smooth_linear}
    \begin{array}{ll}
\displaystyle \minimize_{\mathbf z}         & \displaystyle  ( \nabla f(\mathbf x_0 + \mathbf z^k) )^T(\mathbf z - \mathbf z^k) + \frac{\eta_k}{2} \| \mathbf z - \mathbf z^k \|_2^2 +\frac{\rho}{2} \|    \mathbf z - \mathbf a^\prime \|_2^2 \\
\st & z_i = 0, ~\text{if}~ i \in \mathcal S_\sigma,
    \end{array}
\end{align}
where $\eta_k$ is a decaying parameter associated with the Bregman divergence term 
$\| \mathbf z - \mathbf z^k \|_2^2$.
In problems   \eqref{eq: prob_z_smooth_retrain} and \eqref{eq: prob_z_smooth_linear}, only variables $\{ z_i \}$ satisfying 
 $i \notin \mathcal S_\sigma$ are unknown. The solution to problem \eqref{eq: prob_z_smooth_linear} is then given by
 \begin{align}
\displaystyle [\mathbf z^{k+1}]_i = \left \{
     \begin{array}{ll}
        0  &  i \in \mathcal S_\sigma \\
       \frac{ \eta_k [\mathbf z^k]_i + \rho [\mathbf a^\prime]_i - [\nabla f(\mathbf z^k + \mathbf x_0)]_i }{\eta_k +   \rho }   & i \notin \mathcal S_\sigma.
     \end{array}
     \right.
 \end{align}
 \hfill $\square$

 \section{\textcolor{black}{Adversarial Saliency Map (ASM) and Class Activation Mapping (CAM)}}\label{sec: ASM}
$\mathrm{ASM}(\mathbf x, t) \in \mathbb R^d$ is defined by
 the forward derivative of a neural network given the input sample $\mathbf x$ and the target label $t$ \citep{papernot2016limitations}
\begin{equation}
\label{eq:saliency-map-increasing-features}
\mathrm{ASM}(\mathbf x, t)[i] = \left\lbrace
\begin{array}{ll}
0  & \mbox{ if }   \frac{\partial Z(\textbf{x})_{t}}{\partial \textbf{x}_i}<0  \mbox{ or } \sum_{j\neq t} \frac{\partial Z(\textbf{x})_{j}}{\partial \textbf{x}_i}>0\\
\left(  \frac{\partial Z(\textbf{x})_{t}}{\partial \textbf{x}_i}\right)  \left| \sum_{j\neq t} \frac{\partial Z(\textbf{x})_{j}}{\partial \textbf{x}_i}\right|   & \mbox{ otherwise,}
\end{array}\right.
\end{equation}
where $Z(\mathbf x)_j$ is the $j$th element of logits $Z(\mathbf x)$, representing the output before the last softmax layer in DNNs.
If there exist many classes in a  dataset (e.g., $1000$ classes in ImageNet), then computing $ \sum_{j\neq t} \frac{\partial Z(\textbf{x})_{j}}{\partial \textbf{x}_i}$ is intensive. To circumvent the scalability issue of ASM,  we focus on the logit change with respect to the true label $t_0$ and the target label $t$ only. 
More specifically, we consider three quantities, $\frac{\partial Z(\textbf{x})_{t}}{\partial \textbf{x}_i}$, $-\frac{\partial Z(\textbf{x})_{0}}{\partial \textbf{x}_i}$, and $\left(\frac{\partial Z(\textbf{x})_{t}}{\partial \textbf{x}_i}\right)  \left| \sum_{j\neq t} \frac{\partial Z(\textbf{x})_{j}}{\partial \textbf{x}_i}\right|$,  which correspond to   a) promotion of the score of the target label $t$, b) suppression of the classification score of the true label $t_0$,
and c) a dual role on  suppression and promotion. As a result, we modify 
\eqref{eq:saliency-map-increasing-features} as
\begin{equation}
\label{eq:saliency-map-increasing-features_v2}
\mathrm{ASM}(\mathbf x, t)[i] = \left\lbrace
\begin{array}{ll}
0  & \mbox{ if }   \frac{\partial Z(\textbf{x})_{t}}{\partial \textbf{x}_i}<0  \mbox{ or } \frac{\partial Z(\textbf{x})_{t_0}}{\partial \textbf{x}_i}>0\\
\left(  \frac{\partial Z(\textbf{x})_{t}}{\partial \textbf{x}_i}\right)  \left|  \frac{\partial Z(\textbf{x})_{t_0}}{\partial \textbf{x}_i}\right|   & \mbox{ otherwise.}
\end{array}\right.
\end{equation}



\begin{figure}[htb]  
\centering
\includegraphics[width=1\textwidth]{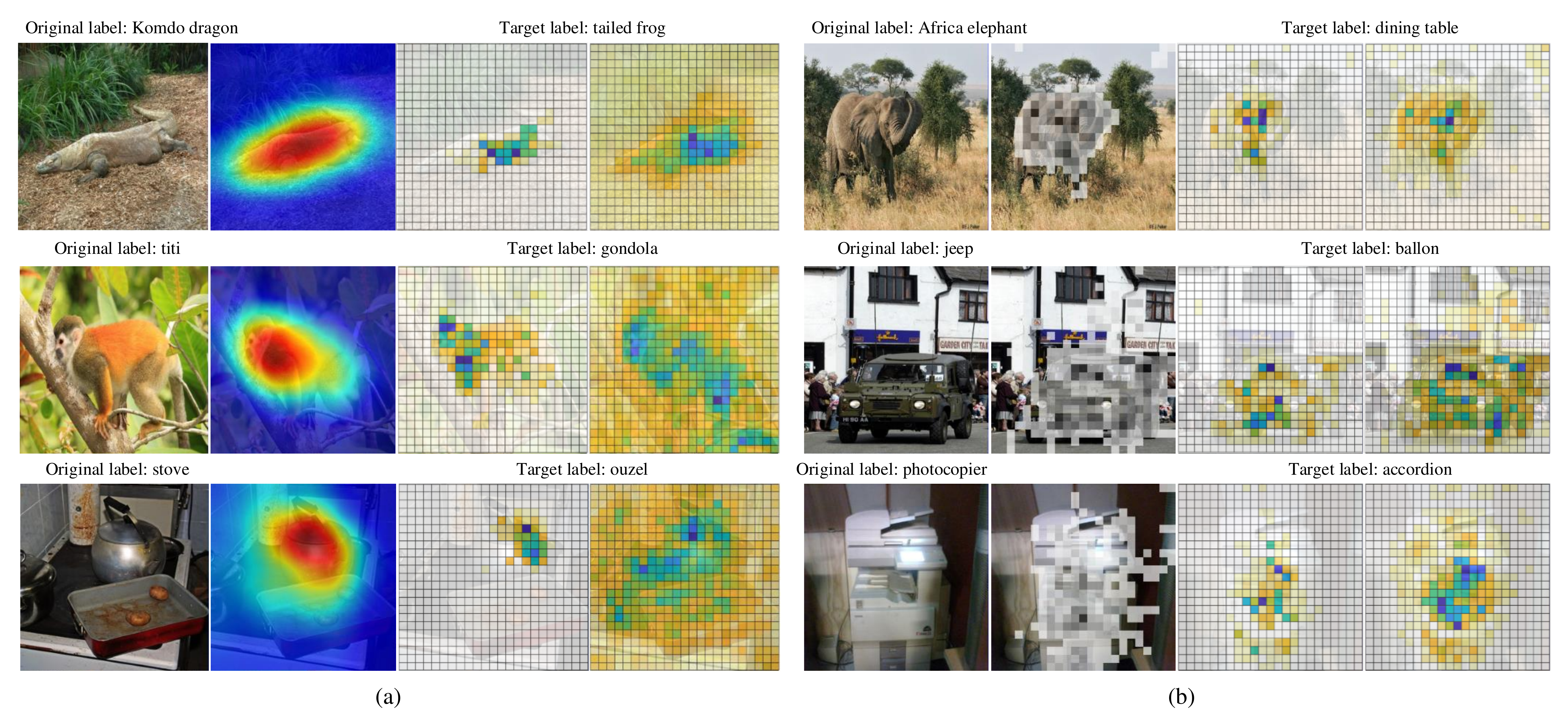}
\caption{
(a) Overlay ASM and $\mathbf B_{\mathrm{ASM}} \circ  \boldsymbol{\delta}$ on top of image with the true and the target label. 
From left to right: original image, ASM (darker color represents larger value of ASM score),  $\mathbf B_{\mathrm{ASM}} \circ  \boldsymbol{\delta}$ under our attack, and $\mathbf B_{\mathrm{ASM}} \circ  \boldsymbol{\delta}$ under C\&W attack.
Here $\nu$ in $\mathbf B_{\mathrm{ASM}}$ is set by the $90$th percentile of ASM scores. 
(b) From left to right: original image, CAM of original label, and perturbations with target label generated from  the StrAttack and C\&W attack, respectively.
}
\label{fig: camsamples}
\end{figure}

 CAM allows us to visualize
the perturbation of adversaries on predicted class scores given  any pair of image and object label, and highlights the discriminative object regions detected by CNNs \citep{zhou2016learning}.
\textcolor{black}{In Fig.\,\ref{fig: camsamples},   
 we show ASM and the discriminative regions identified by CAM on several ImageNet samples.}
 
\section{Experiment Setup and Parameter Setting}  \label{parameter_setting}
\textcolor{black}{In this work, we consider targeted adversarial attacks since they are believed stronger than untargeted attacks. For targeted attacks, we have different methods  to choose the target labels. The average case selects  the  target  label  randomly  among  all  the labels that are not the correct label. The best case performs attacks using all incorrect labels, and report the target label that is the least difficult to attack. The worst case performs attacks using all incorrect labels, and report the target label which is the most difficult to attack.}

In our experiments, two networks are trained for MNIST and CIFAR-10, respectively, and a pre-trained network is utilized for   ImageNet. 
The model architectures for MNIST and CIFAR-10 are the same, both with four convolutional layers, two max pooling layers, two fully connected layers and a softmax layer. 
It can achieve 99.5\% and 80\% accuracy on MNIST and CIFAR-10, respectively. 
For ImageNet,  a pre-trained Inception v3 network \citep{Szegedy2016RethinkingTI} is applied which can achieve 96\% top-5 accuracy.  All experiments are conducted on machines with NVIDIA GTX 1080 TI GPUs.

The implementations of FGM and IFGM are based on the CleverHans package \citep{papernot2016cleverhans}. The key distortion parameter $\epsilon $ is determined by a fine-grained grid search. 
For IFGM, we perform 10 FGM iterations and the distortion parameter $\epsilon'$ is set to $\epsilon/10$ for effectiveness as shown in \citet{tram2018ensemble}.
The implementation of the C\&W attack is based on the opensource code provided by \citet{carlini2017towards}. The maximum iteration number is set to 1000 and it has 9 binary search steps.

In the StrAttack, the group size for MNIST and CIFAR-10 is $2 \times 2$ and its stride is set to 2 if the non-overlapping mask is used, otherwise the group size is $3 \times 3$ and stride is 2. The group size for ImageNet is $13 \times 13$ and its stride is set to 13. 
\textcolor{black}{In ADMM,}
the parameter $\rho$ achieves a trade-off between the convergence rate and the convergence value. A larger $\rho$ could make ADMM converging faster but usually leads to perturbations with larger $\ell_p$ distortion values. A proper configuration of the parameters is suggested as follows:
We set the penalty parameter $\rho = 1$, decaying parameter in \eqref{eq: prob_theta_complete_linear}  $\eta_1= 5$, $\tau = 2$ and $\gamma  = 1$. Moreover, we  set $c $ defined in \eqref{eq: fx}  to   $0.5$ 
for MNIST, $ 0.25$ for CIFAR-10, and $ 2.5$ for ImageNet. Refined attack  technique proposed in Sec. \ref{refined} is applied for all experiments, we set $\sigma$ is equal to 3\% quantile value of non-zero perturbation in $\delta^*$. We observe that 73\% of $\delta^*$ can be retrained to a $\sigma$-sparse perturbation successfully which proof the effective of our refined attack step.

 \section{Supplementary Experimental Results}\label{Results}

\begin{figure}[htb]  
\centering
\includegraphics[width=.9\textwidth]{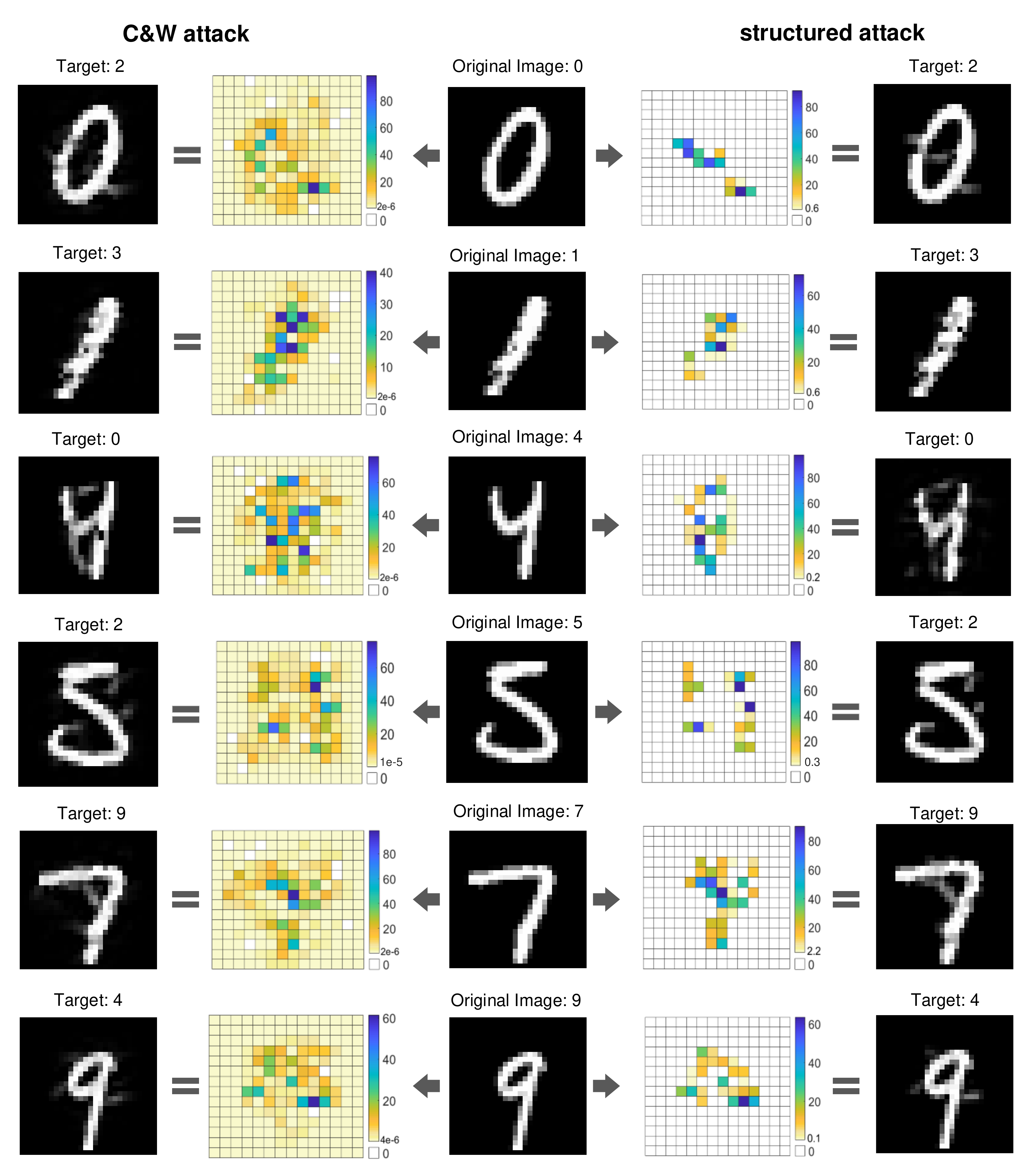}
\caption{C\&W attack vs StrAttack on MNIST with grid size $2 \times 2$.}
\label{fig: more}
\end{figure}

  \begin{figure}[htb]  
  \centering
\includegraphics[width=.9\textwidth]{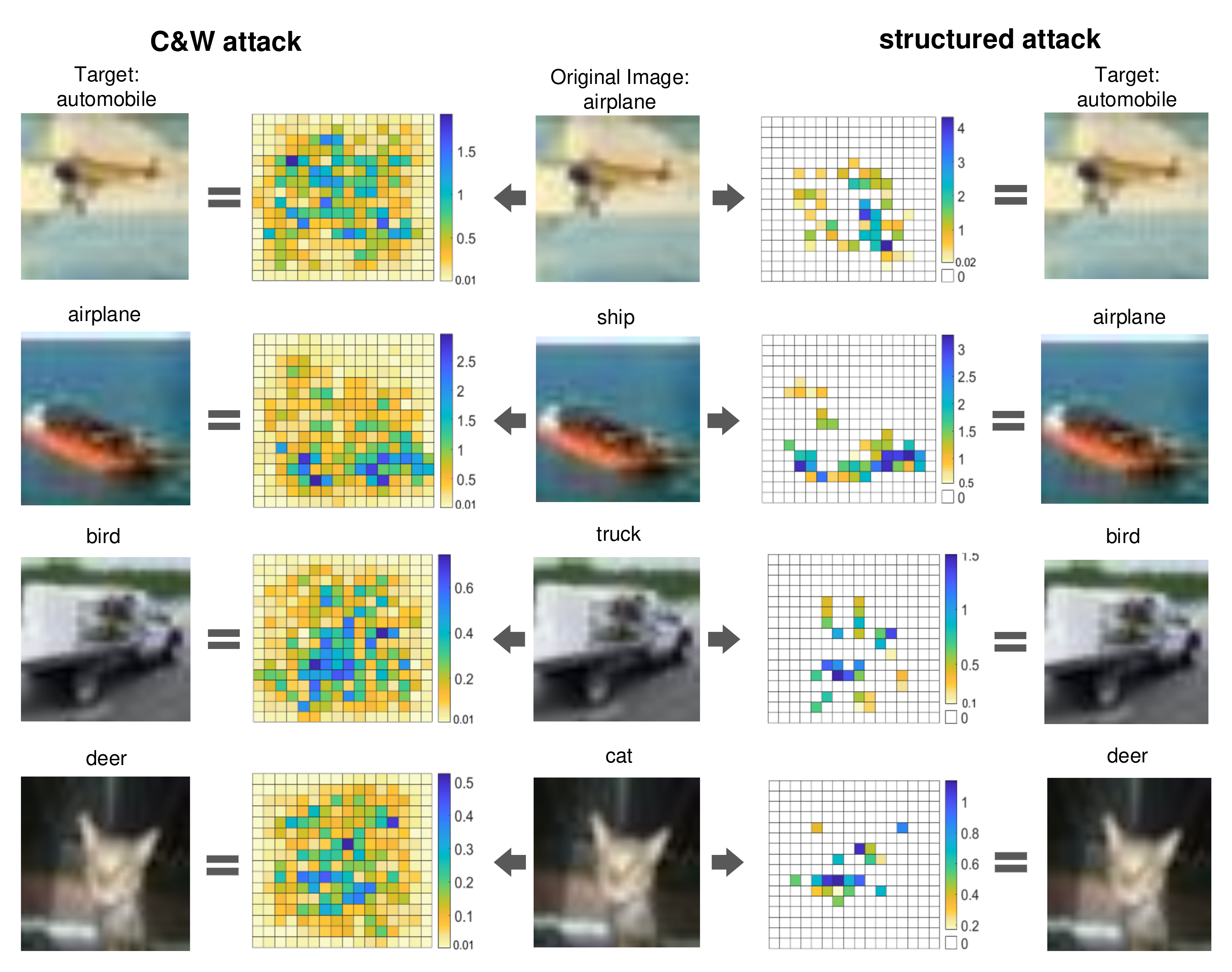}
\caption{C\&W attack vs StrAttack on CIFAR-10 with grid size $2 \times 2$.}
\label{fig: more2}
\end{figure}

  \begin{figure}[htb]  
  \centering
\includegraphics[width=.9\textwidth]{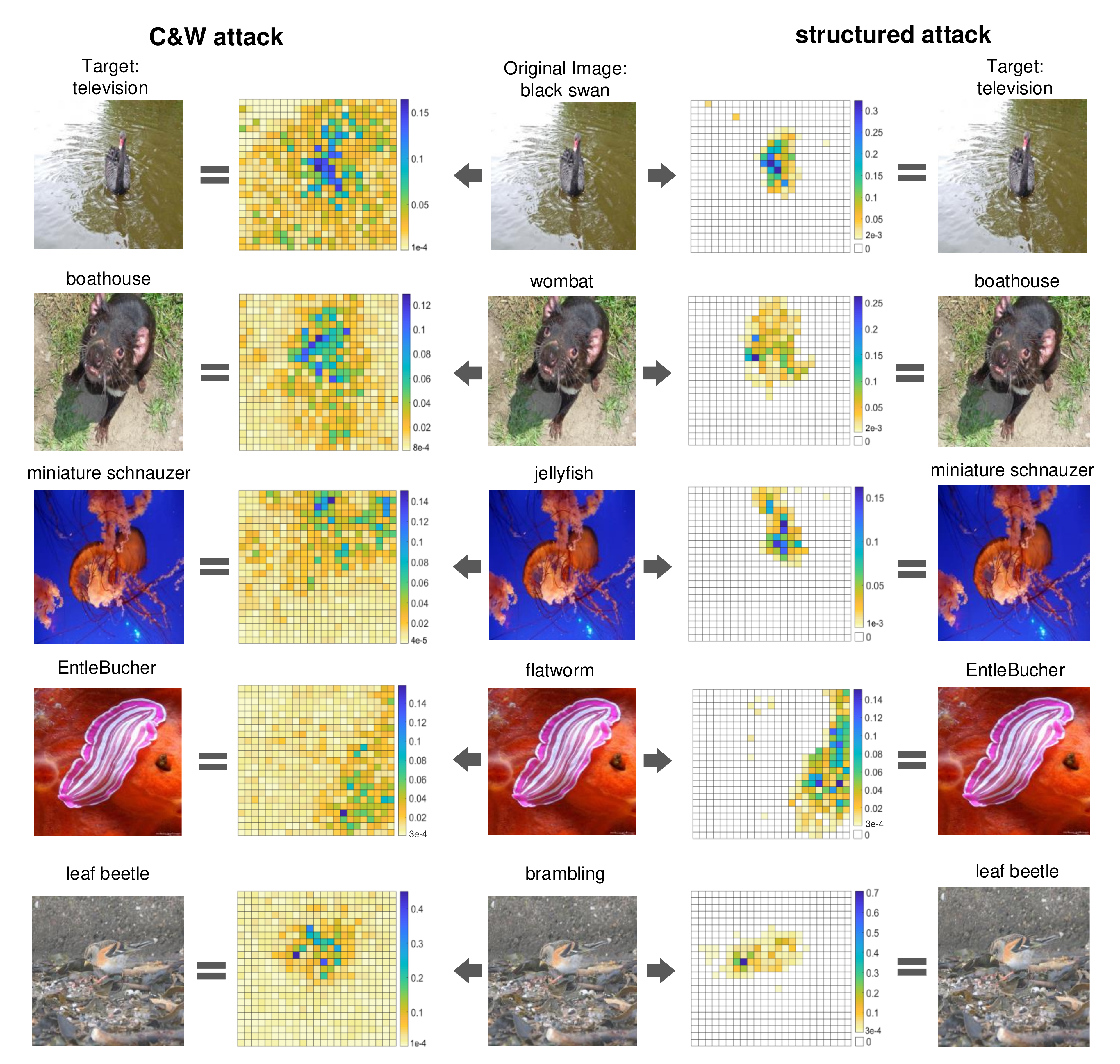}
\caption{C\&W attack vs StrAttack on ImageNet with grid size $13 \times 13$.}
\label{fig: more3}
\end{figure}

Some random choice samples from MNIST (Fig. \ref{fig: more}), CIFAR-10 (Fig. \ref{fig: more2}) and ImageNet (Fig. \ref{fig: more3}) compare StrAttack with C\&W attack. For better sparse visual effect, we only show non-overlapping mask function results here. From these samples, we can discover a consistent phenomenon that our StrAttack is more interested in some particular regions, they usually appear on the objects or their edges in original images, distinctly seen in  MNIST (Fig. \ref{fig: more}) and ImageNet (Fig. \ref{fig: more3}).


\section{StrAttack Against Defensive Distillation and Adversarial Training}  \label{two_defense}
In this section, we present the performance of the StrAttack against  defensive distillation \citep{papernot2016distillation} and adversarial training \citep{tram2018ensemble}.
In defensive distillation, we \textcolor{black}{evaluate} 
the StrAttack for different temperature parameters on MNIST and CIFAR-10. We generate 9000 adversarial examples with 1000 randomly selected images 
\textcolor{black}{from}
MNIST and CIFAR-10, respectively.  The attack success rates of the StrAttack for different temperatures $T$ 
are all 100\%. The reason is that  distillation at temperature $T$ makes the logits approximately $T$ times larger but does not change the relative values of logits. The StrAttack which works on the relative values of logits does not fail.

We further use the StrAttack to break DNNs training on adversarial examples   \citep{tram2018ensemble} with their correct labels on MNIST. The StrAttack is performed on three neural networks: the first network is unprotected, the second is obtained by retraining with 9000 C\&W adversarial examples, and the third network is retained with 9000 adversarial examples crafted by the StrAttack.
The success rate and distortions on the three networks are shown in Table \ref{advertraining}.
The StrAttack can break all three networks with 100\% success rate. 
However, adversarial training shows certain defense effects as an increase on the $\ell_1$ or $\ell_2$ distortion on the latter two networks over the unprotected network is observed.

\begin{table}[htb]  \small
 \centering
  \caption{StrAttack against adversarial training on MNIST } 
  \label{advertraining}
  \scalebox{0.88}[0.88]{
   \begin{threeparttable}
\begin{tabular}{c|ccc|ccc|ccc}
\hline
\makecell{Adversarial} & \multicolumn{3}{c|}{Best case} & \multicolumn{3}{c|}{Average case} & \multicolumn{3}{c}{Worst case} \\
\cline{2-10}
 \makecell{training} & \multicolumn{1}{c}{ASR} & \multicolumn{1}{c}{$\ell_1$} & \multicolumn{1}{c|}{$\ell_2$}  & \multicolumn{1}{c}{ASR} & \multicolumn{1}{c}{$\ell_1$} & \multicolumn{1}{c|}{$\ell_2$} & \multicolumn{1}{c}{ASR} & \multicolumn{1}{c}{$\ell_1$} & \multicolumn{1}{c}{$\ell_2$} \\
\hline
 None &  100 & 10.9 & 1.51 & 100 & 18.05  &2.16& 100&26.9&2.81   \\
  C\&W  & 100 & 16.1 & 1.87 &100 & 25.1 & 2.58 & 100 & 34.2 & 3.26\\
 structured &100& 15.6& 1.86 & 100 & 25.1 & 2.61 & 100 & 34.6 & 3.31\\
\hline
  \end{tabular}
\end{threeparttable}
}
\end{table}

\end{document}